\documentclass[journal,twoside]{IEEEtran}
\IEEEoverridecommandlockouts
\usepackage{color}
\usepackage[x11names]{xcolor}
\usepackage{tikz}
\usetikzlibrary{shapes,arrows,arrows.meta}
\usepackage{mathtools}
\usepackage{amssymb}
\usepackage{lipsum}
\usepackage{float}
\usepackage{makecell}
\usepackage{longtable}
\usepackage{stfloats}
\usepackage{multirow}
\usepackage{amsmath}
\usepackage{bm}
\usepackage{algpseudocode}
\usepackage{booktabs}
\usepackage[numbers,sort&compress]{natbib}
\usepackage{balance}
\usepackage{soul}
\algdef{SE}[SUBALG]{Indent}{EndIndent}{}{\algorithmicend\ }%
\algtext*{Indent}
\algtext*{EndIndent}
\usepackage{xcolor}
\usepackage{threeparttable}
\usepackage{natbib}
\usepackage[breaklinks,citecolor=blue,urlcolor=blue,bookmarks=false,hypertexnames=true]{hyperref}
\allowdisplaybreaks
\usepackage{graphicx}
\usepackage{subfigure}
\usepackage[table]{xcolor}
\usepackage{hhline}
\usepackage[linesnumbered,ruled,vlined]{algorithm2e}

\usepackage{placeins}

\usepackage{soul}
\soulregister\cite7
\soulregister\ref7
\soulregister\pageref7

\begin{document}

\title{OptiPMB: Enhancing 3D Multi-Object Tracking with Optimized Poisson Multi-Bernoulli Filtering}

\author{\IEEEauthorblockN{
Guanhua Ding,~\IEEEmembership{Student Member,~IEEE,}
Yuxuan Xia,~\IEEEmembership{Member,~IEEE,}
Runwei Guan,~\IEEEmembership{Member,~IEEE,}\\
Qinchen Wu,~\IEEEmembership{Student Member,~IEEE,}
Tao Huang,~\IEEEmembership{Senior Member,~IEEE,}
Weiping Ding,~\IEEEmembership{Senior Member,~IEEE,}\\
Jinping Sun,~\IEEEmembership{Member,~IEEE,}
and Guoqiang Mao,~\IEEEmembership{Fellow,~IEEE}
}
\vspace{-0.5cm}

% This work has been submitted to the IEEE for possible publication. Copyright may be transferred without notice, after which this version may no longer be accessible. 
\thanks{© 2025 IEEE. Personal use of this material is permitted. Permission from IEEE must be obtained for all other uses, in any current or future media, including reprinting/republishing this material for advertising or 
promotional purposes, creating new collective works, for resale or redistribution to servers or lists, or reuse of any copyrighted component of this work in other works. Digital Object Identifier \href{https://ieeexplore.ieee.org/document/11218799}{10.1109/TITS.2025.3619674}}
\thanks{The work of Guanhua Ding, Qinchen Wu, and Jinping Sun was supported by National Natural Science Foundation of China, Grant 62131001 and Grant 62171029. \textit{(Corresponding author: Jinping Sun.)}}
\thanks{Guanhua Ding, Qinchen Wu, and Jinping Sun are with the School of Electronic Information Engineering, Beihang University, Beijing 100191, China (e-mail: \url{buaadgh@buaa.edu.cn}; \url{wuqinchen@buaa.edu.cn}; \url{sunjinping@buaa.edu.cn}).}
\thanks{Yuxuan Xia is with the School of Automation and Intelligent Sensing, Shanghai Jiaotong University, Shanghai 200240, China (e-mail: \url{yuxuan.xia@sjtu.edu.cn}).}
\thanks{Runwei Guan is with the Department of Computer Science and Engineering, The Hong Kong University of Science and Technology, Guangzhou 511455, China (e-mail: \url{runwayrwguan@hkust-gz.edu.cn}).}
\thanks{Tao Huang is with the College of Science and Engineering, James Cook University, Cairns QLD 4870, Australia. (e-mail: \url{tao.huang1@jcu.edu.au}).}
\thanks{
Weiping Ding is with the School of Artificial Intelligence and Computer Science, Nantong University, Nantong 226019, China, and also with the Faculty of Data Science, City University of Macau, Macau 999078, China (e-mail: \url{dwp9988@163.com}).}
\thanks{Guoqiang Mao is with Research Laboratory of Smart Driving and Intelligent Transportation Systems, Southeast University, Nanjing 210096, China (e-mail: \url{g.mao@ieee.org}).}
}

% The paper headers
% \markboth{IEEE Transactions on Intelligent Transportation Systems}
% {Ding \MakeLowercase{\textit{et al.}}: OptiPMB: Enhancing 3D Multi-Object Tracking with Optimized Poisson Multi-Bernoulli Filtering}

\maketitle

%%%%%%%%% ABSTRACT
\begin{abstract}
Accurate 3D multi-object tracking (MOT) is crucial for autonomous driving, as it enables robust perception, navigation, and planning in complex environments. While deep learning-based solutions have demonstrated impressive 3D MOT performance, model-based approaches remain appealing for their simplicity, interpretability, and data efficiency. Conventional model-based trackers typically rely on random vector-based Bayesian filters within the tracking-by-detection (TBD) framework but face limitations due to heuristic data association and track management schemes. In contrast, random finite set (RFS)-based Bayesian filtering handles object birth, survival, and death in a theoretically sound manner, facilitating interpretability and parameter tuning. In this paper, we present OptiPMB, a novel RFS-based 3D MOT method that employs an optimized Poisson multi-Bernoulli (PMB) filter while incorporating several key innovative designs within the TBD framework. Specifically, we propose a measurement-driven hybrid adaptive birth model for improved track initialization, employ adaptive detection probability parameters to effectively maintain tracks for occluded objects, and optimize density pruning and track extraction modules to further enhance overall tracking performance. Extensive evaluations on nuScenes and KITTI datasets show that OptiPMB achieves superior tracking accuracy compared with state-of-the-art methods, thereby establishing a new benchmark for model-based 3D MOT and offering valuable insights for future research on RFS-based trackers in autonomous driving.

\end{abstract}

%%%%%%%%% BODY TEXT

\section{Introduction}
\label{Introduction}

%-------------------------------------------------------------------------
\IEEEPARstart{A}{ccurate} and reliable 3D multi-object tracking is essential for autonomous driving systems to enable robust perception, navigation, and planning in complex dynamic environments. Although deep learning has recently driven the development of many learning-based tracking methods \cite{OGR3MOT,CAMO_MOT,ShaSTA,3DMOTFormer,Minkowski,MUTR3D, Motiontrack}, model-based approaches continue to attract significant attention due to their simplicity and data sample efficiency. In model-based tracking, a common design paradigm is to employ random vector (RV)-based Bayesian filtering within the tracking-by-detection (TBD) framework \cite{AB3DMOT,EagerMOT,SimpleTrack, bytetrackv2, DeepFusionMOT, Poly_MOT, Fast_Poly, MCTrack,EMMS-MOT}. Specifically, pre-trained detectors provide object bounding boxes, which serve as inputs for computing similarity scores between detections and existing objects to form an affinity matrix. Assignment algorithms, such as greedy matching and the Hungarian algorithm \cite{NN_GNN}, then establish detection-to-object associations. Bayesian filters (e.g., the Kalman filter) finally update the object state vectors according to the association result. However, because RV-based Bayesian filters estimate each object's state vector individually, these methods often depend on complex heuristic data association steps and track management schemes to effectively track multiple objects. For example, while methods in \cite{EagerMOT, bytetrackv2, DeepFusionMOT, Poly_MOT, Fast_Poly, MCTrack, EMMS-MOT} achieve state-of-the-art performance on large-scale autonomous driving datasets like nuScenes \cite{nuScenes} and KITTI \cite{KITTI}, they employ multi-stage data association and counter/score-based track management to address the inherent limitation of RV-based Bayesian filters.

An important alternative design paradigm within the TBD framework is to utilize random finite set (RFS)-based Bayesian filtering \cite{IV2018PMBM,RFS_M3,GNN_PMB, radar_camera_3D_MOT_GMPHD_1, radar_3D_MOT_GMLMB}. Unlike conventional RV-based filters, RFS filters model the MOT problem within a unified Bayesian framework that naturally handles birth, survival, and death of multiple objects. This comprehensive modeling not only improves the interpretability of the algorithm but also facilitates effective parameter tuning. The Poisson multi-Bernoulli mixture (PMBM) filter was first employed for 3D MOT in autonomous driving scenarios \cite{IV2018PMBM,RFS_M3} due to its elegant handling of detected and undetected objects. Subsequent work \cite{GNN_PMB} proposed a Poisson multi-Bernoulli (PMB) filter utilizing the global nearest neighbor (GNN) data association strategy as an effective approximation of the PMBM filter, which improves computational efficiency and simplifies parameter tuning.
However, existing PMBM/PMB-based trackers \cite{IV2018PMBM,RFS_M3,GNN_PMB} with standard modeling assumptions may still experience performance degradation in complex environments. This motivates the need for innovative algorithm designs tailored for practical autonomous driving scenarios that pushes the limits of RFS-based 3D MOT methods.

To this end, a novel OptiPMB tracker is proposed in this paper. The main contributions are summarized as follows:
\begin{itemize}
\item  We provide a systematic analysis of the RFS-based 3D MOT framework and introduce the OptiPMB tracker, which achieves superior performance in tracking accuracy and track ID maintenance compared with previous RFS-based trackers in autonomous driving scenarios.
\item Within the OptiPMB tracker, we propose a novel hybrid adaptive birth model for effective track initialization in complex environments. Additionally, we employ adaptive detection probability parameters to enhance track maintenance for occluded objects, and optimize the density pruning and track extraction modules to further boost tracking performance.
\item The OptiPMB tracker is comprehensively evaluated and compared with other advanced 3D MOT methods on the nuScenes \cite{nuScenes} and KITTI \cite{KITTI} datasets. Our method achieves state-of-the-art performance on the nuScenes tracking challenge leaderboard with 0.767 AMOTA score, which establishes a new benchmark for model-based online 3D MOT and motivates future research on RFS-based trackers in autonomous driving.
\end{itemize}

The rest of the paper is organized as follows. Related works are reviewed in Section \ref{Related_Works}. The basic concepts of RFS-based 3D MOT are introduced in Section \ref{System Modeling}. Details of our proposed OptiPMB tracker are illustrated in Section \ref{Proposed OptiPMB Tracker}. We evaluate and analyze the performance of OptiPMB on nuScenes and KITTI datasets in Section \ref{Evaluations}. Finally, Section \ref{Conclusion} concludes the paper.

\begin{figure}[t]
\centerline{\includegraphics[width=0.5\textwidth]{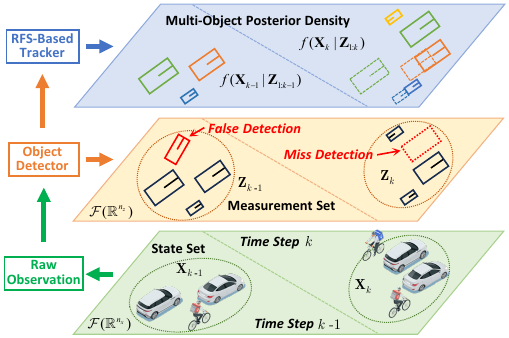}}
\caption{Illustration of the RFS-based 3D MOT system model. The top section displays bounding boxes in different colors, representing tracks with distinct IDs estimated by the RFS-based tracker.}
\label{Fig1}
\end{figure}

%------------------------------------------------------------------------
\section{Related Works}
\label{Related_Works}
\subsection{Model-Based 3D Multi-Object Tracking Methods}

\subsubsection{TBD with RV-Based Bayesian filter}
\label{TBD with RV Bayesian filter}
AB3DMOT \cite{AB3DMOT} establishes an early open-source baseline for 3D MOT on autonomous driving datasets by adopting the TBD framework, which performs tracking with object bounding boxes estimated by pre-trained object detectors. It employs the Hungarian algorithm \cite{NN_GNN} for detection-to-object association and the Kalman filter (KF) to estimate object state vectors, thereby forming a simple model-based paradigm that integrates TBD with RV-based Bayesian filters. Building upon this paradigm, subsequent works have enhanced tracking performance by introducing multi-stage data association \cite{SimpleTrack, bytetrackv2, Poly_MOT, Fast_Poly, MCTrack,EMMS-MOT}, exploiting diverse features to measure affinities between detections and objects \cite{PC3T, ACK3DMOT}, and incorporating multiple sensing modalities \cite{DeepFusionMOT, EagerMOT, radar_lidar_Fusion_Racing, radar_camera_3D_MOT_FARFusion, Feng, StrongFusionMOT, EMMS-MOT}. For example, MCTrack \cite{MCTrack} performs two-stage matching on both the bird-eye's-view (BEV) plane and the range view plane, thereby reducing false positives and ID switch errors when using low-quality detections. EMMS-MOT \cite{EMMS-MOT} leverages spatial synergies between 3D and 2D detection boxes by jointly estimating 3D and 2D motion states with KF and adopting a three-stage data association strategy. Moreover, alternative association methods, including joint probabilistic data association \cite{Jrmot_JPDA, radar_camera_3D_MOT_1_JPDA} and multiple hypothesis tracking \cite{MHT_3DMOT}, have also been explored within this paradigm.

\subsubsection{TBD with RFS-Based Bayesian filter}
\label{TBD with RFS Bayesian filter}
Unlike conventional RV-based Bayesian filters, RFS-based filters model the collection of object states as a set-valued random variable, providing a mathematically rigorous framework to handle uncertainties in object existence, births, and deaths. An early attempt to utilize an RFS-based Bayesian filter within the TBD framework for 3D MOT is proposed in \cite{IV2018PMBM}, where the PMBM filter is used to estimate object trajectories from monocular camera detections. RFS-M$^3$ \cite{RFS_M3} further extends the PMBM filter to LiDAR-based 3D MOT by using 3D bounding boxes as inputs. 
% proposing a TBD pipeline that utilizes 3D bounding box detections as inputs. 
To improve efficiency and simplify parameter tuning, \cite{GNN_PMB} combines the PMB filter with GNN data association, proposing a simple yet effective GNN-PMB tracker. Other RFS-based Bayesian filters, such as the probability hypothesis density and labeled multi-Bernoulli filters, are also explored for 3D MOT applications \cite{radar_3D_MOT_GMLMB, radar_camera_3D_MOT_GMPHD_1}.

\subsubsection{JDT with MEOT}
\label{JDT with EOT}
While the TBD framework relies on object detectors to provide inputs for trackers, multi-extended object tracking (MEOT) methods under the joint-detection-and-tracking (JDT) framework can simultaneously estimate the motion, shape, and size of objects directly from raw point cloud measurements, reducing the reliance on accurate 3D bounding box detections. Different statistical modeling of geometrically extended objects are utilized by RFS-based filters to achieve 3D MEOT in \cite{LiDAR_3D_Extended_Target_MOT, LiDAR_3D_Extended_Target_MOT_2, radar_3D_Extended_Target_MOT}. Additionally, camera images are used to improve point cloud clustering quality and data association accuracy for 3D MEOT in \cite{camera_radar_3D_Extended_Target_MOT_2, camera_radar_3D_Extended_Target_MOT_3}.

\subsection{Learning-Based 3D Multi-Object Tracking Methods}

\subsubsection{TBD with Learning-Assisted Data Association}
\label{Learning Embedding for TBD with Bayesian filter}
The classical strategy to enhance model-based 3D MOT with neural networks is employing learning-assisted data association.
In this approach, feature embeddings captured by neural networks serve as an additional cue to construct the affinity matrix, which is subsequently processed by conventional data association methods, such as greedy matching \cite{Probabilistic3DMM, OGR3MOT, radar_camera_3D_MOT_DL_3, ShaSTA, LEGO} and multi-stage Hungarian algorithm \cite{CAMO_MOT, C_and_L_KITTI_Latest_TITS, DINO-MOT}. Object states are then estimated by the Kalman filter based on association results. For example, DINO-MOT \cite{DINO-MOT} leverages the DINOv2 encoder to extract visual features of pedestrians and refines the association results by computing cosine similarities between embeddings. Similar strategies are also applied to RFS-based filters \cite{NEBP, NEBP-V3}, where learned embeddings complement the belief propagation messages to enhance the probabilistic data association.

\subsubsection{Learning-Based JDT}
\label{Learning Embedding for JDT with RV Bayesian filter}
Recent advances in learning-based JDT methods leverage neural networks to simultaneously detect objects and establish associations across frames. For instance, GNN3DMOT \cite{Gnn3dmot} employs a graph neural network that fuses 2D and 3D features to capture complex spatial relationships, while Minkowski Tracker \cite{Minkowski} utilizes sparse spatial-temporal convolutions to efficiently perform JDT on point clouds. Other approaches combine cross-modal cues from cameras and LiDAR \cite{AlphaTrack, JMODT} or incorporate probabilistic and geometric information \cite{PF_MOT,PolarMOT}. Transformer-based methods \cite{TransMOT, Trajectoryformer, 3DMOTFormer} further enhance performance by capturing long-range dependencies through attention mechanisms. These learning-based JDT approaches often offer improved accuracy and robustness in complex autonomous driving scenarios, but non-learnable assignment algorithms are still required to obtain the hard association results.

\subsection{End-to-End Learnable 3D Multi-Object Tracking Methods}

\subsubsection{
Learnable Data Association Module}
Although neural networks have been utilized in learning-based 3D MOT, the non-differentiable assignment modules could hinder fully data-driven training of the tracker and subsequent networks. An intuitive solution is using learnable modules for data association. For instance, RaTrack \cite{RaTrack} replaces the Hungarian algorithm with a differentiable alternative, proposing an end-to-end trainable tracker for 4D radars. SimTrack \cite{SimTrack} eliminates the heuristic matching step by utilizing a hybrid-time centerness map to handle object birth, death, and association.

\subsubsection{Tracking-by-Attention}
The tracking-by-attention strategy was initially proposed for 2D MOT \cite{MOTR, Trackformer} and has been extended to the 3D domain in recent studies \cite{MUTR3D, PF-Track, Motiontrack}. The core idea is to represent objects with dedicated variable queries learned from data. In each frame, new birth queries with unique IDs are generated and subsequently propagated as existing object queries, thus enabling the network to implicitly handle data association with label assignment rules during training. This approach facilitates end-to-end optimization and makes such transformer-based trackers a key component in recent autonomous driving pipelines \cite{UniAD, ViP3D}.

%------------------------------------------------------------------------

\section{RFS-Based 3D MOT Revisited}
\label{System Modeling}
\subsection{Basic Concept and Notation}
In this section, we introduce the fundamental concepts and notations of RFS-based 3D MOT. For a 3D object, the state vector $\mathbf{x}\in \mathbb{R}^{n_x}$ usually consists of motion states (e.g., position, velocity, acceleration, turn rate) and extent states (e.g., shape, size, and orientation of the object bounding box).
Under the RFS framework illustrated in Fig. \ref{Fig1}, the multi-object state is defined by a finite set $\mathbf{X}\in \mathcal{F}(\mathbb{R}^{n_x})$, where $\mathcal{F}(\mathbb{R}^{n_x})$ is the space of all finite subsets of the object state space $\mathbb{R}^{n_x}$. Assuming that $N_k$ objects exist at time step $k$, then the multi-object state can be represented as $\mathbf{X}_k=\{\mathbf{x}_k^1,\mathbf{x}_k^2,...,\mathbf{x}_k^{N_k}\}$, whose cardinality (the number of elements) is $|\mathbf{X}_k|=N_k$. 

In autonomous driving applications, raw observations of the objects (e.g., camera images, LiDAR and radar point clouds) are often preprocessed by detectors to obtain bounding boxes, enabling the trackers to perform MOT using these detections under the TBD framework. 
Since TBD-based 3D MOT methods were proven to be simple and effective in many previous studies, we employ the TBD framework in this paper and assume that the measurement of an object is a bounding box detection. At time step $k$, assume that the objects are observed by a set of measurements $\mathbf{Z}_k=\{\mathbf{z}_k^1,\mathbf{z}_k^2,...,\mathbf{z}_k^{M_k}\}\in \mathcal{F}(\mathbb{R}^{n_z})$. Each measurement $\mathbf{z}_k\in \mathbb{R}^{n_z}$ is a bounding box detected from raw observations which contains measured object states such as position, velocity, size, and orientation. The collection of measurement sets from time step $1$ to $k$ is denoted by $\mathbf{Z}_{1:k}$.In practical scenarios, some objects can be misdetected, and the set of measurements $\mathbf{Z}$ often consists of not only true object detections but also false detections (clutter), as shown in Fig. \ref{Fig1}. The ambiguous relation between the objects and the measurements increases the complexity of 3D MOT, while the RFS-based Bayesian filtering aims to provide an effective and rigorous solution to the multi-object state estimation problem.

\subsection{Key Random Processes}
\label{Key_Random_Process}
There are three important RFSs widely used in MOT system modeling, which are the Poisson point process (PPP) RFS, the Bernoulli RFS, and the multi-Bernoulli RFS. A PPP RFS $\mathbf{X}$ has Poisson-distributed cardinality and independent, identically distributed elements, which can be defined by the probability density function
\begin{equation}
    f^{\mathrm{ppp}}(\mathbf{X})=e^{-\mu}\prod_{\mathbf{x}\in\mathbf{X}}\mu p(\mathbf{x})=e^{-\int \lambda(\mathbf{x})\mathrm{d}\mathbf{x}}\prod_{\mathbf{x}\in\mathbf{X}}\lambda(\mathbf{x}).
    \label{PPP}
\end{equation}
Here, $\mu$ is the Poisson rate of cardinality, $p(\mathbf{x})$ is the spatial distribution of each element, and the intensity function $\lambda(\mathbf{x})=\mu p(\mathbf{x})$ completely parametrizes $\mathbf{X}$. Due to its simplicity, the PPP RFS is often used to model undetected objects, newborn objects, and false detections in RFS-based trackers.

A Bernoulli RFS $\mathbf{X}$ contains a single element with probability $r$ or is empty with probability $1-r$. The probability density function of $\mathbf{X}$ is given by
\begin{equation}
    f^{\mathrm{ber}}(\mathbf{X})=
    \begin{cases}
        1-r&\mathbf{X}=\emptyset\\
        rp(\mathbf{x})&\mathbf{X}=\{\mathbf{x}\}\\
        0&|\mathbf{X}|\geq2.
    \end{cases}
    \label{Ber}
\end{equation}
The Bernoulli RFS can be used to model the state distribution and the measurement likelihood function of an object.

A multi-Bernoulli RFS $\mathbf{X}$ is a union of $N$ independent Bernoulli RFSs $\mathbf{X}^{i}$, i.e., $\mathbf{X}=\bigcup_{i=1}^N\mathbf{X}^{i}$, whose probability density function is defined by
\begin{equation}
    f^{\mathrm{mb}}(\mathbf{X})=
    \begin{cases}
    \sum_{\uplus_{i\in\mathbf{I}}\mathbf{X}^{i}=\mathbf{X}}\prod_{i\in\mathbf{I}}f^{\mathrm{ber}}(\mathbf{X}^{i})\hspace{-0.5em}&|\mathbf{X}|\leq N\\
    0&|\mathbf{X}|>N\\
    \end{cases}
    \label{MB}
\end{equation}
where $\mathbf{I}=\{1,...,N\}$ is an index set for the Bernoulli RFSs, and $\uplus_{i\in\mathbf{I}}\mathbf{X}^{i}=\mathbf{X}$ denotes that $\mathbf{X}$ is the union of mutually disjoint subsets $\{\mathbf{X}^{i}\}$. The multi-Bernoulli RFS can be applied for modeling states of multiple objects.

\subsection{Bayesian Filtering with Conjugate Prior Densities}
\label{Bayesian_Filtering_with_Conjugate_Prior_Densities}
The goal of the RFS-based Bayesian filter is to recursively estimate the multi-object posterior $f(\mathbf{X}_k|\mathbf{Z}_{1:k})$, from which the state of each object can be extracted. A Bayesian filtering recursion includes a prediction step followed by an update step. In the prediction step, the posterior density in the last time step is predicted to the current time step by the Chapman-Kolmogorov equation \cite{point_target_MOT_PMBM}
\begin{equation}
    \resizebox{.9\hsize}{!}{$f(\mathbf{X}_{k}|\mathbf{Z}_{1:k-1})=\int{g(\mathbf{X}_{k}|\mathbf{X}_{k-1})f(\mathbf{X}_{k-1}|\mathbf{Z}_{1:k-1})\delta\mathbf{X}_{k-1}}$}
\end{equation}
where $g(\mathbf{X}_{k}|\mathbf{X}_{k-1})$ is the multi-object state transition density.
Under conventional MOT assumptions of the RFS framework \cite{point_target_MOT_PMBM,point_target_MOT_GLMB,GNN_PMB,point_target_MOT_PMB_BP}, an object with state $\mathbf{x}_{k-1}$ survives from time step $k-1$ to $k$ with probability $p_\mathrm{s}(\mathbf{x}_{k-1})$, and its state transits with $g(\mathbf{x}_k|\mathbf{x}_{k-1})$. The newborn objects are modeled by a PPP RFS with intensity $\lambda^\mathrm{b}(\cdot)$.

In the update step, the predicted density $f(\mathbf{X}_{k}|\mathbf{Z}_{1:k-1})$ is updated using the information of measurements $\mathbf{Z}_k$. Given the multi-object measurement likelihood $h(\mathbf{Z}_k|\mathbf{X}_k)$, the multi-object posterior can be calculated by Bayes' rule \cite{point_target_MOT_PMBM}
\begin{equation}
    f(\mathbf{X}_k|\mathbf{Z}_{1:k})=\frac{h(\mathbf{Z}_k|\mathbf{X}_k)f(\mathbf{X}_k|\mathbf{Z}_{1:k-1})}{\int{h(\mathbf{Z}_k|\mathbf{X}_k)f(\mathbf{X}_k|\mathbf{Z}_{1:k-1})\delta\mathbf{X}}}.
\end{equation}
Here, the physical process of objects generating measurements is modeled in the likelihood function $h(\mathbf{Z}_k|\mathbf{X}_k)$. Based on the conventional assumptions, each object is detected with probability $p_\mathrm{d}(\mathbf{x}_k)$, and a detected object generates a measurement according to the single-object measurement likelihood $h(\mathbf{z}_k|\mathbf{x}_k)$. The false detections (clutter) are assumed to follow a PPP RFS with intensity $\lambda^\mathrm{c}(\cdot)$.

Conjugate prior densities are essential for efficient Bayesian filtering, as such densities can preserve the same mathematical form throughout filtering recursions. PMBM \cite{point_target_MOT_PMBM} and $\delta$-generalized labeled multi-Bernoulli ($\delta$-GLMB) \cite{point_target_MOT_GLMB} are two multi-object conjugate prior densities widely used in RFS-based Bayesian filters. In this research, we adopt a simplified form of the PMBM density, known as the PMB density, for use in our proposed OptiPMB tracker.

\begin{figure*}[t]
\centering
\includegraphics[width=1.0\textwidth]{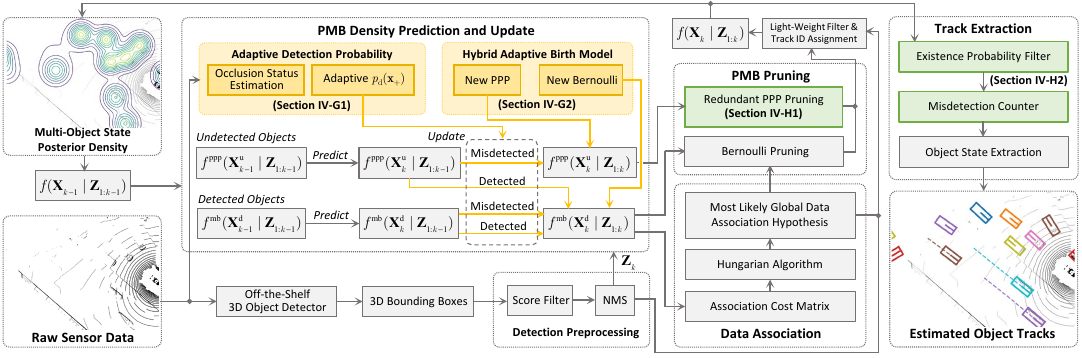}
\caption{Overall pipeline of the proposed OptiPMB tracker. Compared to the previous state-of-the-art PMB filter-based 3D MOT method \cite{GNN_PMB}, the innovative differences and improvements of OptiPMB are highlighted in distinct colors. \textcolor{DarkGoldenrod2}{\textbf{Yellow}} denotes adaptive designs to improve the robustness of the tracker (see Section \ref{Adaptive Designs} for details). \textcolor{DarkOliveGreen4}{\textbf{Green}} denotes algorithm modules optimized for better tracking performance (see Section \ref{Optimizations} for details).}
\label{Fig2}
\vspace{-0.2cm}
\end{figure*}

\section{OptiPMB Tracker: Improving the Performance of RFS-Based 3D MOT}
\label{Proposed OptiPMB Tracker}
In this section, we provide a detailed illustration of our OptiPMB tracker and explain how its innovative designs and modules enhance the performance of RFS-based 3D MOT. The overall pipeline of the OptiPMB tracker is depicted in Fig. \ref{Fig2}, with a comprehensive description of each component presented below.

\subsection{Single-Object State Transition and Measurement Model}
\label{single_object_transition_and_measurement_model}
Before delving into the 3D MOT pipeline, it is necessary to define the object state transition and measurement models. As an RFS-based tracker, OptiPMB represents the multi-object state as a set of single-object state vectors, $\mathbf{X}=\{\mathbf{x}^i\}_{i\in\mathbf{I}}$, where each $\mathbf{x}\in\mathbf{X}$ is defined as:
\begin{equation}
\begin{aligned}
    \mathbf{x}&=[\mathbf{x}_\mathrm{M}^\mathrm{T},\mathbf{x}_\mathrm{A}^\mathrm{T},\mathbf{x}_\mathrm{U}^\mathrm{T},x_{\mathrm{cnt}},x_\mathrm{len},x_s]^\mathrm{T}
\end{aligned}
\end{equation}
where $\mathbf{x}_\mathrm{M}=[x,y,v,\phi,\omega,a]^\mathrm{T}$, $\mathbf{x}_\mathrm{A}=[\alpha,\beta,\gamma,\zeta]^\mathrm{T}$, and $\mathbf{x}_\mathrm{U}=[c,t]^\mathrm{T}$.
The motion state $\mathbf{x}_\mathrm{M}$ comprises the object's position on the BEV plane $(x,y)$, velocity $v$, heading angle $\phi$, turn rate $\omega$, and acceleration $a$. The auxiliary state $\mathbf{x}_\mathrm{A}$ includes the length $\alpha$, width $\beta$, and height $\gamma$ of the object bounding box as well as $\zeta$, the object's position along the $z$-axis. The time-invariant state $\mathbf{x}_\mathrm{U}$ contains the object class $c$ and track ID $t$. Finally, $x_{\mathrm{cnt}}$ records the number of consequent misdetections for this object, $x_{\mathrm{len}}$ records the number of time steps the object has survived, $x_s$ is a confidence score utilized in certain evaluation metrics. It is worth noting that only the motion state $\mathbf{x}_\mathrm{M}$ is estimated by Bayesian filtering recursions, while the non-motion states $(\mathbf{x}_\mathrm{A}, \mathbf{x}_\mathrm{U}, x_{\mathrm{cnt}}, x_{\mathrm{len}}, x_s)$ are processed by a light-weight filter (see Section \ref{Light-Weight Filter for Non-Motion States} for details). 

Accurate modeling of object motion is crucial for achieving high tracking performance. In OptiPMB, the constant turn rate and acceleration (CTRA) model \cite{CTRA} is employed to predict the object's motion state, formulated by
\begin{equation}
    \mathbf{x}_{\mathrm{M},{k}}=\Gamma_\mathrm{CTRA}(\mathbf{x}_{\mathrm{M},k-1})+\mathbf{v}_{k}
\end{equation}
where $\Gamma_\mathrm{CTRA}(\cdot)$ denotes the nonlinear dynamic function of the CTRA model (see \cite[Section III-A]{CTRA}), $\mathbf{v}_{k}\sim\mathcal{N}(\mathbf{0},Q_{k})$ is Gaussian-distributed process noise. Unscented transform (UT) \cite{UKF} is applied to handle the nonlinear functions, and we have the motion state transition density
\begin{equation}
\begin{aligned}
    g(\mathbf{x}_{\mathrm{M},k}|\mathbf{x}_{\mathrm{M},k-1}) &= \mathcal{N}(\mathbf{x}_{\mathrm{M},k};\mathbf{m}_{k|k-1},P_{k|k-1})\\
    P_{k|k-1}&=\bar{P}_{k|k-1} + Q_{k}\\
    (\mathbf{m}_{k|k-1},\bar{P}_{k|k-1})&=\mathrm{UT}[\Gamma_\mathrm{CTRA}(\cdot),\mathbf{m}_{k-1},P_{k-1}].
\end{aligned}
\label{Motion Predict}
\end{equation}
Here, $\mathbf{m}_{k-1}$ and $P_{k-1}$ denote the posterior mean and covariance of the motion state, i.e., $\mathbf{x}_{\mathrm{M},k-1}\sim\mathcal{N}(\mathbf{m}_{k-1},P_{k-1})$. The transformation $\mathrm{UT}[\Gamma(\cdot),\mathbf{m},P]$ computes the mean and covariance projected through the nonlinear function $\Gamma(\cdot)$. The non-motion states remain unchanged in the transition model.
The survival probability is simplified as a predefined constant dependent only on the object's class, i.e., $p_\mathrm{s}(c)$.

In the TBD framework, removing false detections generated by the object detector helps reduce false tracks and improves computational efficiency \cite{SimpleTrack, GNN_PMB}. To achieve this, OptiPMB employs a per-class score filter and non-maximum suppression (NMS) for detection preprocessing \cite{GNN_PMB}. Specifically, bounding boxes with detection scores below $\eta_\mathrm{sf}$\label{eta_sf} are removed, and NMS is applied for overlapping boxes with intersection over union (IoU) \cite{GIOU} exceeding $\eta_\mathrm{iou}$\label{eta_iou}. After preprocessing, the remaining detections form the measurement set $\mathbf{Z}=\{\mathbf{z}^m\}_{m\in\mathbf{M}}$, where each measurement $\mathbf{z}\in\mathbf{Z}$ is defined by 

\begin{equation}
\begin{aligned}
    \mathbf{z}&=[\mathbf{z}_\mathrm{M}^\mathrm{T},\mathbf{z}_\mathrm{A}^\mathrm{T},z_c,z_s]^\mathrm{T}
\end{aligned}
\end{equation}
where $\mathbf{z}_\mathrm{M}=[z_x,z_y,z_{vx},z_{vy},z_\phi]^\mathrm{T}$ and $\mathbf{z}_\mathrm{A}=[z_\alpha,z_\beta,z_\gamma,z_\zeta]^\mathrm{T}$.
Here, $\mathbf{z}_\mathrm{M}$ is the observation of the object’s motion state, while $\mathbf{z}_\mathrm{A}$ denotes the observation of the object’s auxiliary state. The motion measurement model is given by
\begin{equation}
\label{Single Object Measurement Model}
    \begin{aligned}
        \mathbf{z}_{\mathrm{M},k}&= \mathrm{H}(\mathbf{x}_{\mathrm{M},k})+\mathbf{w}_k \\
        &= [x_k, y_k, v_k\cos\phi_k, v_k\sin\phi_k, \phi_k]^\mathrm{T} +\mathbf{w}_k
    \end{aligned}
\end{equation}
where $\mathbf{w}_{k}\sim\mathcal{N}(\mathbf{0},R_{k})$ is Gaussian measurement noise. Since the object's non-motion states are not estimated by Bayesian filtering, we simply assume $\mathbf{z}_{\mathrm{A},k}=\mathbf{x}_{\mathrm{A},k}$ and $z_{c,k}=c_k$. The detection score of the measurement $\mathbf{z}_{k}$ is denoted by $z_{s,k}\in(0,1]$, which is provided by the 3D object detector.

\subsection{Basic Framework: Bayesian Filtering with PMB Density}
\label{Basic_Framework_Bayesian_Filtering}
The OptiPMB tracker is built upon the basic framework of a PMB filter, which represents the multi-object state with a PMB RFS and propagates its probability density through the Bayesian filtering recursions outlined in Section \ref{Bayesian_Filtering_with_Conjugate_Prior_Densities}. Unlike the PMBM filter \cite{point_target_MOT_PMBM}, which maintains multiple probable global data association hypotheses\footnote{The discussion on global and local hypotheses could be found in the following content of this sub-section.}, the PMB filter selects and propagates only the best global hypothesis. Consequently, when properly designed and parameterized, the PMB filter can achieve higher computational efficiency without compromising 3D MOT performance \cite{GNN_PMB}.

The OptiPMB tracker defines the multi-object posterior density as follows
\begin{equation}
    f(\mathbf{X}_k|\mathbf{Z}_{1:k})=\ \ \sum_{\mathclap{\mathbf{X}^\mathrm{u}_k\uplus\mathbf{X}^\mathrm{d}_k=\mathbf{X}_k}}\ \ f^\mathrm{ppp}(\mathbf{X}_k^\mathrm{u}|\mathbf{Z}_{1:k}) f^\mathrm{mb}(\mathbf{X}_k^\mathrm{d}|\mathbf{Z}_{1:k})
    \label{PMB}
\end{equation}
where the multi-object state is modeled as a union of two disjoint subsets, i.e., $\mathbf{X}^\mathrm{u}_k\uplus\mathbf{X}^\mathrm{d}_k=\mathbf{X}_k$. The PPP RFS $\mathbf{X}^\mathrm{u}_k$ denotes the state of \textit{potential objects which have never been detected}, while the MB RFS $\mathbf{X}^\mathrm{d}_k$ denotes the state of \textit{potential objects which have been detected at least once}. This partition of detected and undetected objects enables an efficient hypotheses management of potential objects \cite{point_target_MOT_PMB_BP, point_target_MOT_PMBM}.

According to the definitions in Section \ref{Key_Random_Process}, the posterior PMB density at time step $k-1$, $f(\mathbf{X}_{k-1}|\mathbf{Z}_{1:k-1})$, can be fully determined by a set of parameters
\begin{equation}
\label{PMB parameters}
    \lambda^\mathrm{u}_{k-1},\{(r_{k-1}^{i},p_{k-1}^{i})\}_{i\in\mathbf{I}_{k-1}}.
\end{equation}
Here, $\lambda_{k-1}^\mathrm{u}(\cdot)$ is the intensity of the PPP RFS $\mathbf{X}^\mathrm{u}_{k-1}$. Based on our system modeling in Section \ref{single_object_transition_and_measurement_model}, the PPP intensity $\lambda_{k-1}^\mathrm{u}(\cdot)$ is a weighted mixture of Gaussian-distributed Poisson components, defined as
\begin{equation}
    \lambda_{k-1}^\mathrm{u}(\mathbf{x})= \textstyle\sum_{n=1}^{N_{k-1}^\mathrm{u}} \mu^{\mathrm{u},n}\mathcal{N}(\mathbf{x};\mathbf{m}_{k-1}^{\mathrm{u},n},P_{k-1}^{\mathrm{u},n})
    \label{PPP Posterior}
\end{equation}
where $\mu^{\mathrm{u},n}$, $\mathbf{m}_{k-1}^{\mathrm{u},n}$, and $P_{k-1}^{\mathrm{u},n}$ represent the weight, motion state mean, and motion state covariance of the $n$-th Poisson component, respectively.
The MB RFS $\mathbf{X}^\mathrm{d}_{k-1}$ consists of a set of Bernoulli components, represented by $\{(r_{k-1}^{i},p_{k-1}^{i})\}_{i\in\mathbf{I}_{k-1}}$ in \eqref{PMB parameters}. The $i$-th Bernoulli component, $\mathbf{X}^i_{k-1}$, is characterized by existence probability $r_{k-1}^i$ and Gaussian spatial distribution $p_{k-1}^{i}\sim\mathcal{N}(\mathbf{m}_{k-1}^i,P_{k-1}^i)$. To improve the clarity of hypotheses management, OptiPMB employs the track-oriented hypothesis structure \cite{point_target_MOT_PMBM, GNN_PMB}, defining the MB index set as
\begin{equation}
    \mathbf{I}_{k-1}=\textstyle\bigcup_{n=1}^{N^\mathrm{d}_{k-1}}\mathbf{L}_{k-1}^n
\end{equation}
where $N^\mathrm{d}_{k-1}$ denotes the number of previously detected potential objects, and $\mathbf{L}^n_{k-1}$ is the index set for the local data association hypotheses of the $n$-th potential object. 

\begin{figure}[t]
\centerline{\includegraphics[width=0.5\textwidth]{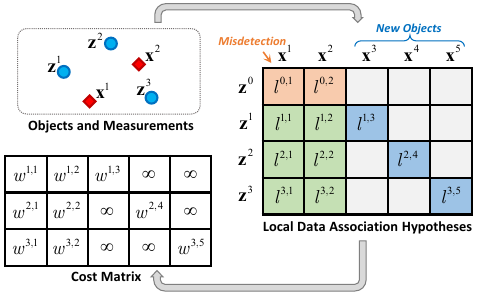}}
\caption{Cost matrix and local data association hypotheses. The misdetection, detection, and first-time detection hypotheses are highlighted in \textcolor{orange}{orange}, \textcolor{DarkOliveGreen4}{green}, and \textcolor{blue}{blue}, respectively. Cost matrix entries with infinity values represent impossible association hypotheses.}
\label{Fig3}
\vspace{-0.2cm}
\end{figure}

A \textbf{local hypothesis} represents the scenario where an object either generates a measurement or is misdetected at the current time step. As illustrated in Fig. \ref{Fig3}, consider a case with two previously detected potential objects, $\{\mathbf{x}^1,\mathbf{x}^2\}$, and three measurements, $\{\mathbf{z}^1,\mathbf{z}^2,\mathbf{z}^3\}$. According to the assumptions and system modeling in Section \ref{System Modeling} and \ref{single_object_transition_and_measurement_model}, three types of local hypotheses can then be identified:
\begin{itemize}
\item \textbf{Misdetection Hypotheses}: $\{l^{0,1},l^{0,2}\}$, corresponding to objects $\{\mathbf{x}^1,\mathbf{x}^2\}$ being misdetected (i.e., associated with a dummy measurement $\mathbf{z}^0$).
\item \textbf{Detection Hypotheses}: $\{l^{m,n}|n=1,2;m=1,2,3\}$, corresponding to cases where a previously detected potential object is detected at the current time step and generates one of the measurements.
\item \textbf{First-Time Detection Hypotheses}:$\{l^{1,3},l^{2,4},l^{3,5}\}$, indicating that the measurements originate from newly detected objects or clutter, corresponding to new potential objects $\{\mathbf{x}^3,\mathbf{x}^4,\mathbf{x}^5\}$.
\end{itemize}
These three types of local hypotheses cover all possible data associations between objects and measurements. A \textbf{global hypothesis} consists of a set of local hypotheses that are compatible with the system modeling, defining a valid association between all objects and measurements. For example, $\{l^{0,1},l^{2,2},l^{1,3},l^{3,5}\}$ is a valid global hypothesis, while $\{l^{1,1},l^{1,2},l^{3,2}\}$ is invalid, because it violates the constraint that a measurement must associate with one and only one potential object. Since OptiPMB only maintains the best global data association hypothesis over time, each detected potential object retains a single local hypothesis that is included in the global hypothesis, while all other local hypotheses are pruned after the PMB update. Therefore, the local hypothesis index set of the $n$-th detected potential object is defined as $\mathbf{L}_{k-1}^n=\{n\}$. Details on local hypothesis management are in Section \ref{PMB_Density_Update}. 

As illustrated in Fig. \ref{Fig2}, at the subsequent time step $k$, the PMB density $f(\mathbf{X}_{k-1}|\mathbf{Z}_{1:k-1})$ is first propagated through the prediction step and then updated using the preprocessed object detections. The situation that previously undetected potential objects remain undetected at the current time step is also considered within this pipeline (see Section \ref{PPP_Update}) alongside the three types of local hypotheses. Following the PMB prediction and update, the data association module determines the optimal global data association hypothesis, which is then utilized as protocol for the pruning module to eliminate redundant PPP and Bernoulli components from the PMB density. Next, the non-motion states (including track IDs) of detected potential objects are determined by a light-weight filter. Finally, the estimated posterior $f(\mathbf{X}_k|\mathbf{Z}_{1:k})$ is processed by the track extraction module to generate the object tracks. The details of our OptiPMB tracker are explained in the following subsections, where the time subscript $k|k-1$ is abbreviated as $+$ for simplicity. Pseudo code for the complete OptiPMB recursion is provided in Algorithm \ref{Alg1}.

\renewcommand{\thealgocf}{1}
\begin{algorithm}[htbp]
\footnotesize
\caption{Pseudo Code of OptiPMB}
\label{Alg1}

\KwIn{Parameters of the PMB posterior at the previous time step $\lambda^\mathrm{u}_{k-1},\{(r^{i}_{k-1},p^{i}_{k-1})\}_{i\in\mathbf{I}_{k-1}}$, LiDAR point cloud, and detected object 3D bounding boxes at the current time step.}

\KwOut{Estimated object tracks at current time step $k$.}

\textit{\textbf{Note:} The subscript $+$ is the abbreviation of $k|k-1$. All detection probability $p_\mathrm{d}(\mathbf{x}_+)$ are adaptively determined (see Section \ref{Adaptive Filtering Parameters}).}

\textit{/* PMB prediction (see Section \ref{PMB_Density_Prediction}) */} \\

\For(\textit{// PPP prediction}){$n\in\{1,...,N_{k-1}^\mathrm{u}\}$}{
    Compute $\mathbf{m}_{+}^{\mathrm{u},n},P_{+}^{\mathrm{u},n}$ from $\mathbf{m}_{k-1}^{\mathrm{u},n},P_{k-1}^{\mathrm{u},n}$ using \eqref{Motion Predict}\;
}
Compute $\lambda_{+}^\mathrm{u}$ from $\mathbf{m}_{+}^{\mathrm{u},n},P_{+}^{\mathrm{u},n},\mu_{k-1}^{\mathrm{u},n}$ using \eqref{PPP_prediction}\;

\For(\textit{// MB prediction}){$i\in\mathbf{I}_{k-1}$}{
    Compute $\mathbf{m}_{+}^{i},P_{+}^{i}$ from $\mathbf{m}_{k-1}^{i},P_{k-1}^{i}$ using \eqref{Motion Predict}\;
    Compute $r_+^{i},p^i_+$ from $\mathbf{m}_{+}^{i},P_{+}^{i}$ using \eqref{MB_prediction}\;
}

\textit{/* PMB update (see Section \ref{PMB_Density_Update}) */} \\

Preprocess detected object 3D bounding boxes, obtain the measurement set $\mathbf{Z}_k$\;

\For(\textit{// Iterate over previously detected potential objects}){$n\in \{1,...,N^\mathrm{d}_{k-1}\}$}{
    $\hat{\mathbf{L}}_k^n\gets \{(0,n)\}$; \textit{// Create misdetection hypothesis}\\
    Compute $r_k^{0,n}, p_k^{0,n}$ from $r_+^{n},p_+^n$ using \eqref{Misdetect Hypothesis Update}\;
    Perform gating with $r_+^{n},p_+^n$, obtain the gate area for the $n$-th potential object\;
    \For{$\mathbf{z}_k^m\in\mathbf{Z}_k$}{
        \If{$\mathbf{z}_k^m$ is in the gate}{
            $\hat{\mathbf{L}}_k^n\gets \hat{\mathbf{L}}_k^n\cup \{(m,n)\}$; \textit{// Create detection hypothesis}\\
            Compute $r_k^{m,n}, p_k^{m,n}, w_k^{m,n}$ from $r_+^{n},p_+^n,\mathbf{z}_k^m$ using \eqref{Detection Hypothesis Update}\;
        }
    }
}

Initialize: $\mathbf{M}_k^\mathrm{low}\gets \emptyset$\;

\For(\textit{// Iterate over measurements}){$\mathbf{z}_k^m\in\mathbf{Z}_k$}{
    $n\gets N_{k-1}^\mathrm{d}+m$, $\hat{\mathbf{L}}_k^{n}\gets \{(m,n)\}$; \textit{// Create first-time detection hypothesis}\\
    Obtain the gate area of $\mathbf{z}_k^m$\;

    From $\lambda_{+}^\mathrm{u}$, obtain the index set $\mathbf{J}_k^m$ for Poisson components inside the gate of $\mathbf{z}_k^m$\;

    \If(\textit{// Associate with undetected potential objects}){$\mathbf{J}_k^m\neq\emptyset$}{
        $\lambda \gets \sum_{j\in\mathbf{J}_k^m}\mu^{\mathrm{u},j}p_\mathrm{s}(c^{\mathrm{u},j})\mathcal{N}(\mathbf{x};\mathbf{m}_{+}^{\mathrm{u},j},P_{+}^{\mathrm{u},j})$\;
        \For{$j\in\mathbf{J}_k^m$}{
            Compute $e^j, p^j$ from $\mathbf{z}_k^m,\lambda$ using \eqref{First-Time Detection Hypothesis Update 2}\;
        }
        Compute $r_k^{m,n}, p_k^{m,n}, w_k^{m,n}$ from $\{(e^j,p^j)\}_{j\in\mathbf{J}_k^m}$ using \eqref{First-Time Detection Hypothesis Update 1}\;
    }
    \Else(\textit{// HABM (see Section \ref{Hybrid_Adaptive_Birth_Model})}){
        Compute $p_\mathrm{a}(\mathbf{z}_k^m)$ using \eqref{Association Probability}\;

        \If(\textit{// Low-score measurement}){$z_{s,k}^m<\eta_\mathrm{score}$}{
            Compute $r_k^{m,n}, w_k^{m,n}$ using \eqref{Clutter Hypothesis}\;
            $\mathbf{M}_k^\mathrm{low}\gets \mathbf{M}_k^\mathrm{low} \cup\{m\}$\;
        }
        \Else(\textit{// High-score measurement}){
            Compute $r_k^{m,n}, p_k^{m,n}, w_k^{m,n}$ using \eqref{Newborn Hypothesis}\;
        }
    
}
}

$\hat{N}_k^\mathrm{d}\gets N_{k-1}^\mathrm{d}+|\mathbf{Z}_k|$, $\hat{\mathbf{I}}_k\gets \uplus_{n=1}^{\hat{N}_k^\mathrm{d}} \hat{\mathbf{L}}_k^n$\;
Compute $\lambda_k^{\mathrm{b}}$ from $\{[\mathbf{z}_k^m, p_\mathrm{a}(\mathbf{z}_k^m)]\}_{m\in\mathbf{M}_k^\mathrm{low}}$ using \eqref{HABM}\;
Compute $\hat{\lambda}_k^{\mathrm{u}}$ from $\lambda_+^{\mathrm{u}},\lambda_k^{\mathrm{b}}$ with \eqref{PPP Update}; \textit{// PPP update}\\

\textit{/* Data association and PMB pruning (see Section \ref{Data Association and PMB Pruning} and Section \ref{Redundant PPP Pruning}) */} \\
From $\{w_k^{m,n}\}$, construct the cost matrix $W$\;
Obtain the optimal global hypothesis from $W$ with the Hungarian algorithm\;
Prune redundant Bernoulli and Poisson components from the updated PMB density $\hat{\lambda}^\mathrm{u}_{k},\{(r^{i}_{k},p^{i}_{k})\}_{i\in \hat{\mathbf{I}}_{k}}$\;
\textit{/* Light-weight filtering (see Section \ref{Light-Weight Filter for Non-Motion States}) */} \\
Update non-motion states with the light-weight filter\;
Reconstruct MB and PPP densities with \eqref{Reconstruct MB} and \eqref{Reconstruct PPP}, obtain the posterior PMB density $\lambda^\mathrm{u}_{k},\{(r^{i}_{k},p^{i}_{k})\}_{i\in \mathbf{I}_{k}}$\;

\textit{/* Track extraction (see Section \ref{Track Extraction}) */} \\
From $\lambda^\mathrm{u}_{k},\{(r^{i}_{k},p^{i}_{k})\}_{i\in \mathbf{I}_{k}}$, extract the object tracks at the current time step.
\end{algorithm}

\subsection{PMB Density Prediction}
\label{PMB_Density_Prediction}
In the PMB prediction step, the posterior PPP intensity is predicted as a Gaussian mixture according to the state transition model in Section \ref{single_object_transition_and_measurement_model}. Specifically, we have 
\begin{equation}
\label{PPP_prediction}
\begin{aligned}
    \lambda_{+}^\mathrm{u}(\mathbf{x})&= \textstyle\sum_{n=1}^{N_{k-1}^\mathrm{u}} \mu^{\mathrm{u},n}_+ \mathcal{N}(\mathbf{x};\mathbf{m}_{+}^{\mathrm{u},n},P_{+}^{\mathrm{u},n})\\
    \mu^{\mathrm{u},n}_+ &= \mu^{\mathrm{u},n}_{k-1} p_\mathrm{s}(c^{\mathrm{u},n})
\end{aligned}
\end{equation}
where the predicted mean $\mathbf{m}_{+}^{\mathrm{u},n}$ and covariance $P_{+}^{\mathrm{u},n}$ of the $n$-th Poisson component is obtained from \eqref{Motion Predict}. Note that OptiPMB does not introduce the birth intensity $\lambda^\mathrm{b}(\cdot)$ during the prediction step, as is done in conventional PMBM/PMB implementations \cite{point_target_MOT_PMBM, GNN_PMB, IV2018PMBM, RFS_M3}. Instead, object birth is managed in the update step using the newly proposed adaptive birth model (see Section \ref{Hybrid_Adaptive_Birth_Model} for details). 

The predicted MB density is parameterized by the existence probability and spatial distribution of Bernoulli components, i.e., $\{(r_{+}^i,p_{+}^i)\}_{i\in\mathbf{I}_{+}}$, where
\begin{equation}
\label{MB_prediction}
\begin{cases}
    r_{+}^i=r_{k-1}^ip_\mathrm{s}(c^{i})\\
    p_{+}^i(\mathbf{x})=\mathcal{N}(\mathbf{x};\mathbf{m}_{+}^i,P_{+}^i).
\end{cases}
\end{equation}
The predicted mean $\mathbf{m}_{+}^{i}$ and covariance $P_{+}^{i}$ of the $i$-th Bernoulli component are also calculated by \eqref{Motion Predict}. The MB index set and local hypotheses index sets remain unchanged during the PMB prediction, i.e., $\mathbf{I}_+=\mathbf{I}_{k-1}$, $\{\mathbf{L}_+^n\}=\{\mathbf{L}_{k-1}^n\}$.

\subsection{PMB Density Update}
\label{PMB_Density_Update}
In the PMB update step, OptiPMB enumerates local data association hypotheses (defined in Section \ref{Basic_Framework_Bayesian_Filtering}) and update the predicted PMB density using Bayes' rule. 
Update procedures for each type of local hypothesis are described as follows.

\subsubsection{MB Update for Misdetection and Detection Hypotheses}
\label{Misdetect and Detection Hypotheses}
After PMB prediction, misdetection and detection hypotheses are generated for previously detected potential objects, which are characterized by the predicted MB density $\{(r_{+}^i,p_{+}^i)\}_{i\in\mathbf{I}_{+}}$. As discussed in Section \ref{Basic_Framework_Bayesian_Filtering}, the $n$-th previously detected potential object retains only one local hypothesis indexed by $n$. The corresponding Bernoulli component $\mathbf{X}_{+}^n$ represents the predicted object state with the existence probability $r_{+}^n$ and spatial distribution $p_{+}^{n}(\cdot)$ defined in \eqref{MB_prediction}. A misdetection hypothesis $l_k^{0,n}$ is first generated as \cite{point_target_MOT_PMBM}
\begin{equation}
\label{Misdetect Hypothesis Update}
\begin{cases}
    r=\frac{r_+^n(1-p_\mathrm{d}(\mathbf{x}_+^n))}{1-r_+^n + r_+^n(1-p_\mathrm{d}(\mathbf{x}_+^n))}\\
    p(\mathbf{x})=p_+^n(\mathbf{x})\\
\end{cases}
\end{equation}
where $(r,p)$ are the parameters of the corresponding Bernoulli component. Next, class gating is applied to select measurements belonging to the same class as the object. Distance gating is then performed to identify measurements that fall within a predefined distance $\eta_\mathrm{dist}$ from the object. Each measurement $\mathbf{z}^m_k$ that falls within the gate of the object generates a detection hypothesis $l_k^{m,n}$, defined as \cite{point_target_MOT_PMBM}
\begin{equation}
\label{Detection Hypothesis Update}
\begin{cases}
    w=-\ln \frac{r_+^n p_\mathrm{d}(\mathbf{x}_+^n) \mathcal{N}(\mathbf{z}_{xy,k}^m;\hat{\mathbf{z}}^n_{k},S^n_k)} {1-r_+^n+r_+^n(1- p_\mathrm{d}(\mathbf{x}_+^n))}\\
    r=1\\
    p(\mathbf{x})=\mathcal{N}(\mathbf{x};\mathbf{m}_{k}^{m,n},P_k^{m,n})
\end{cases}
\end{equation}
where $w$ is the association cost, $\mathbf{z}_{xy,k}^m=[z_{x,k}^m,z_{y,k}^m]^\mathrm{T}$ is the position measurement, $\hat{\mathbf{z}}^n_{k}$ and $S^n_k$ denote the mean and covariance of the predicted position measurement.
Real-world data indicates that measured object velocity and heading are often inaccurate and noisy, leading to fluctuations in association costs. To mitigate the impact of outlier measurements, OptiPMB employs a simplified measurement model for computing association costs, which relies solely on position information. The measurement model is formulated as
\begin{equation}
\begin{aligned}
\label{Simplified measurement model}
    \mathbf{z}_{xy,k} &= H_{xy}\mathbf{x}_{\mathrm{M},k} + \mathbf{w}_{xy,k}\\
    H_{xy} &= \begin{bmatrix}
        1 & 0 & 0 & 0 & 0 & 0\\
        0 & 1 & 0 & 0 & 0 & 0
    \end{bmatrix}
\end{aligned}
\end{equation}
where $\mathbf{w}_{xy,k}\sim\mathcal{N}(\mathbf{0},R_{xy,k})$, $R_{xy,k}$ corresponds to the upper-left $2\times2$ sub-matrix of the full covariance matrix $R_k$. Consequently, the mean and covariance of the predicted position measurement in \eqref{Detection Hypothesis Update} are computed by
\begin{equation}
\label{Predicted Measurement}
\begin{cases}
    \hat{\mathbf{z}}_{k}^n = H_{xy} \mathbf{m}_+^n\\
    S_k^n = H_{xy} P_{+}^n H_{xy}^\mathrm{T} +R_{xy,k}.
\end{cases}
\end{equation}
Notably, \eqref{Simplified measurement model} is used solely for calculating association costs, whereas the complete measurement model \eqref{Single Object Measurement Model} is applied in the unscented Kalman filter (UKF) \cite{UKF} to update the object's motion state. With measurement $\mathbf{z}_k^m$, the motion state mean $\mathbf{m}_{k}^{m,n}$ and covariance $P_k^{m,n}$ in \eqref{Detection Hypothesis Update} are updated as
\begin{equation}
\label{UKF Update}
    (\mathbf{m}_{k}^{m,n},P_k^{m,n})=\mathrm{UKF}[\mathrm{H}(\cdot),\mathbf{m}_{+}^{n},P_+^{n},\mathbf{z}_k^m,R_k].
\end{equation}

After generating all possible misdetection and detection local hypotheses for the $n$-th object, the index set of these new local hypotheses is defined by
\begin{equation}
    \hat{\mathbf{L}}_k^n=\{(0,n)\} \cup \{(m,n)|m\in \mathbf{M}^n_k)\},\ \ n=1,...,N_{k-1}^d
\end{equation}
where $(0,n)$ indexes the misdetection hypothesis, $\{(m,n)\}$ denote the indices for detection hypotheses, $\mathbf{M}_k^n$ is the index set for measurements within the gate of the $n$-th object.

\subsubsection{MB Update for First-Time Detection Hypotheses}
\label{First-Time Detection Hypotheses}
For each measurement, a newly detected potential object is created and a first-time detection hypothesis is generated for this new object. Since the PPP components represent undetected potential objects, if at least one Poisson component falls within the gate of a measurement $\mathbf{z}^m_k$, then the first-time detection hypothesis $l_k^{m,N^\mathrm{d}_{k-1}+m}$ is generated as \cite{point_target_MOT_PMBM}
\begin{equation}
\label{First-Time Detection Hypothesis Update 1}
\begin{cases}
    w=-\ln[e+\lambda^\mathrm{c}(\mathbf{z}_k^m)]\\
    r= e/[e+\lambda^\mathrm{c}(\mathbf{z}_k^m)]\\
    p(\mathbf{x})=\frac{1}{e}\sum_{j\in\mathbf{J}^m_k} e^j p^j(\mathbf{x})
\end{cases}
\end{equation}
where
\begin{equation}
\label{First-Time Detection Hypothesis Update 2}
\begin{cases}
    e=\sum_{j\in\mathbf{J}^m_k} e^j\\
    e^j= \mu_{+}^{\mathrm{u},j} p_\mathrm{d}(\mathbf{x}_+^{\mathrm{u},j})\mathcal{N}(\mathbf{z}_{xy,k}^m;\hat{\mathbf{z}}^{\mathrm{u},j}_{k},S_k^{\mathrm{u},j})\\
    p^j(\mathbf{x})=\mathcal{N}(\mathbf{x};\mathbf{m}_{k}^{m,j},P_k^{m,j})
\end{cases}
\end{equation}
and the set $\mathbf{J}^m_k$ denotes the indices of Poisson components that fall inside the gate of $\mathbf{z}_k^m$. The clutter intensity $\lambda^\mathrm{c}(\cdot)$ is characterized by a clutter rate parameter $\mu^\mathrm{c}(\cdot)$ predefined for each object class and a uniform spatial distribution $p_\mathrm{uni}(\cdot)$ over the region of observation, i.e.,
\begin{equation}
    \lambda^\mathrm{c}(\mathbf{z}_k^m)=\mu^\mathrm{c}(z_{c,k}^m) p_\mathrm{uni}(\mathbf{z}_{xy,k}^m)= \mu^\mathrm{c}(z_{c,k}^m)/A
\end{equation}
where $A$ denotes the area size of the observation region. The mean and covariance of the predicted position measurement, $(\hat{\mathbf{z}}^{\mathrm{u},j}_{k},S_k^{\mathrm{u},j})$, as well as the updated mean and covariance of the motion state, $(\mathbf{m}_{k}^{m,j},P_k^{m,j})$, are calculated in a manner similar to \eqref{Predicted Measurement} and \eqref{UKF Update}. The updated density $p(\mathbf{x})$ is approximated as a single Gaussian density via moment matching \cite{point_target_MOT_PMBM}. The detection probability $p_\mathrm{d}(\mathbf{x}_+)$ is adaptively determined based on the object's occlusion status, as further elaborated in Section \ref{Adaptive Filtering Parameters}. If no Poisson component falls inside the measurement's gate, the measurement is processed using the proposed hybrid adaptive birth model (HABM) to determine the hypothesis parameters $(w,r,p)$. The HABM also adaptively generates $\lambda_k^\mathrm{b}(\cdot)$ based on current measurements to model the undetected newborn objects. Details of the HABM are in Section \ref{Hybrid_Adaptive_Birth_Model}.

The local hypothesis index set for each new object is
\begin{equation}
    \hat{\mathbf{L}}_k^n =\{(n-N_{k-1}^\mathrm{d},n)\},\ \ n=N_{k-1}^\mathrm{d}+1,...,N_{k-1}^\mathrm{d}+M_k.
\end{equation}
After enumerating all possible local hypotheses, the index set of the updated MB density is represented as
\begin{equation}
    \hat{\mathbf{I}}_k = \textstyle\bigcup_{n=1}^{\hat{N}^\mathrm{d}_{k}}\hat{\mathbf{L}}_{k}^n
\end{equation}
where the number of detected potential objects increases by the number of measurements, i.e., $\hat{N}^\mathrm{d}_{k}=N^\mathrm{d}_{k-1}+M_k$.

\subsubsection{PPP Update}
\label{PPP_Update}
The updated PPP intensity for undetected potential objects can be simply expressed as \cite{point_target_MOT_PMBM}
\begin{equation}
\label{PPP Update}
\begin{aligned}
    \hat{\lambda}_{k}^\mathrm{u}&(\mathbf{x}) = \textstyle\sum_{n=1}^{\hat{N}_{k}^\mathrm{u}} \mu^{\mathrm{u},n}_k \mathcal{N}(\mathbf{x};\mathbf{m}_{k}^{\mathrm{u},n},P_{k}^{\mathrm{u},n})\\
    &= \textstyle\sum_{n=1}^{N_{k-1}^\mathrm{u}} \mu^{\mathrm{u},n}_+ p_{\mathrm{d}}(\mathbf{x}_+^{\mathrm{u},n}) \mathcal{N}(\mathbf{x};\mathbf{m}_{+}^{\mathrm{u},n},P_{+}^{\mathrm{u},n}) + \lambda_k^\mathrm{b}(\mathbf{x})
\end{aligned}
\end{equation}
since no measurement update is applied to its components.

\subsection{Data Association and PMB Pruning}
\label{Data Association and PMB Pruning}
In the PMB update procedure, the association costs $w$ for each detection and first-time detection local hypothesis are calculated, forming a cost matrix, as shown in Fig. \ref{Fig3}. The optimal global data association hypothesis, which minimizes the total association cost while assigning each measurement to a detected potential object, is determined from this cost matrix using the Hungarian algorithm \cite{point_target_MOT_PMBM, GNN_PMB}. To reduce the computational complexity of data association, OptiPMB employs gating to eliminate unlikely hypotheses and set the corresponding association costs to infinity. After the data association, Bernoulli components that are not included in the optimal global hypothesis or determined as clutter by the HABM are removed to reduce unnecessary filtering computation. If no Bernoulli component belongs to a detected potential object after pruning, the object is then removed. The redundant Poisson components are also pruned to reduce false tracks and ID switch errors. See Section \ref{Redundant PPP Pruning} for details of PPP pruning.

\subsection{Light-Weight Filter for Non-Motion States}
\label{Light-Weight Filter for Non-Motion States}
As discussed in Section \ref{single_object_transition_and_measurement_model}, the non-motion states of an object are processed using a light-weight filter to enhance computational efficiency. Specifically, when a newly detected potential object is created from a measurement $\mathbf{z}_k^m$, the time-invariant state is determined by
\begin{equation}
    \mathbf{x}_\mathrm{U}=[c,t]^\mathrm{T}=[z_{c,k}^m,(k,m)]^\mathrm{T}.
\end{equation}
Here, the track ID $(k,m)$ uniquely identifies a new object with the time step $k$ and the measurement index $m$. Throughout the OptiPMB tracking recursions, the time-invariant states remain unchanged. The counter variables for misdetections and survival time steps are initialized as $x_{\mathrm{cnt},k}=0$ and $x_{\mathrm{len},k}=1$. The object's confidence score is defined as
\begin{equation}
\label{Confidence Score}
    x_{s,k}=[1-\exp(-x_{\mathrm{len},k})]z_{s,k}^m
\end{equation}
where $z_{s,k}^m\in(0,1]$ represents the detection score of $\mathbf{z}_{k}^m$, indicating that $x_{s,k}$ is further down-scaled from the detection score for tracks with shorter survival time.

After the optimal global hypothesis is determined and the PMB density is pruned, if a remaining detected potential object is associated with a measurement $\mathbf{z}_{k}^m$ under the global hypothesis, its auxiliary state is updated by
\begin{equation}
    \mathbf{x}_{\mathrm{A},k}=(1-z_{s,k}^m)\mathbf{x}_{\mathrm{A},k-1} + z_{s,k}^m\mathbf{z}_{\mathrm{A},k}^m,
\end{equation}
the misdetection counter $x_{\mathrm{cnt},k}$ is reset to zero, and the confidence score $x_{s,k}$ is calculated by \eqref{Confidence Score}. Otherwise, if the object is misdetected, the auxiliary state remains unchanged, the counter increments by one $x_{\mathrm{cnt},k}=x_{\mathrm{cnt},k-1}+1$, and $x_{s,k}$ is set to zero. The number of survival time steps increases as $x_{\mathrm{len},k}=x_{\mathrm{len},k-1}+1$ regardless of the association status.

After pruning and light-weight filtering, the MB density is re-indexed as
\begin{equation}
\label{Reconstruct MB}
    \mathbf{I}_k= \textstyle\bigcup_{n=1}^{N^\mathrm{d}_{k}} \mathbf{L}_{k}^n = \textstyle\bigcup_{n=1}^{N^\mathrm{d}_{k}} \{n\}
\end{equation}
where $N_k^\mathrm{d}$ is the number of remaining detected potential objects. The PPP intensity after pruning is expressed as
\begin{equation}
\label{Reconstruct PPP}
    \lambda_k^\mathrm{u} = \textstyle\sum_{n=1}^{N_{k}^\mathrm{u}} \mu^{\mathrm{u},n}_k \mathcal{N}(\mathbf{x};\mathbf{m}_{k}^{\mathrm{u},n},P_{k}^{\mathrm{u},n})
\end{equation}
where $N_k^\mathrm{u}$ denotes the number of remaining Poisson components. The posterior PMB density is then fully determined by the parameters $(\lambda_k^\mathrm{u},\{(r_{k}^i,p_{k}^i)\}_{i\in\mathbf{I}_{k}})$.

% \vspace{-2pt}
\subsection{Adaptive Designs}
\label{Adaptive Designs}
The performance of model-based 3D MOT methods depends heavily on the alignment between the pre-designed model and the actual tracking scenarios. To enhance robustness of the tracker, OptiPMB proposes multiple adaptive modules and integrates them in the Bayesian filtering pipeline, enabling dynamic parameter adjustment and effective initialization of new object tracks.

\subsubsection{\textbf{Adaptive Detection Probability (ADP)}}
\label{Adaptive Filtering Parameters}
In urban traffic scenarios, objects are frequently occluded by the environment and other objects, leading to inaccurate detections or even complete misdetections. To improve the track continuity and reduce ID switch errors, OptiPMB adaptively adjusts the detection probability based on the object's occlusion status, which is estimated using the LiDAR point cloud and the object 3D bounding box. Specifically, during the PMB update procedure, the predicted object 3D bounding boxes are projected onto the LiDAR coordinates, and the number of LiDAR points within each bounding box is counted. The adaptive detection probability is then defined as
\begin{equation}
    p_\mathrm{d}(\mathbf{x}_+)=p_\mathrm{d0}(c)\cdot\min[1,(1-s_\mathrm{d}(c))\frac{\mathrm{PTS(\mathbf{x}_+)}}{\mathrm{PTS}_0(c)}+s_\mathrm{d}(c)]
\end{equation}
where the baseline detection probability $p_\mathrm{d0}(c)$, minimal scaling factor $s_\mathrm{d}(c)\in(0,1]$, and expected number of LiDAR points $\mathrm{PTS}_0(c)$ are parameters predefined based on the object's class $c$, $\mathrm{PTS(\mathbf{x}_+)}$ counts the actual number of LiDAR points within the object's predicted bounding box. According to this definition, the detection probability decreases as $\mathrm{PTS(\mathbf{x}_+)}$ becomes smaller, indicating that the object may be occluded. This adaptive mechanism enable OptiPMB to maintain tracks for occluded objects and reduce ID switch errors, improving robustness in complex urban environments.

\subsubsection{\textbf{Hybrid Adaptive Birth Model (HABM)}}
\label{Hybrid_Adaptive_Birth_Model}
The design of the object birth model not only affects the initiation of object trajectories but also influences the association between existing objects and measurements. Consequently, it plays a critical role in the performance of RFS-based tracking algorithms. OptiPMB employs a hybrid adaptive birth model that integrates the conventional PPP birth model with a measurement-driven adaptive birth method, enabling fast-response and accurate initialization of new object tracks. 

According to the PMB update procedure described in Section \ref{First-Time Detection Hypotheses}, measurements that cannot be associated with any Poisson component are processed by the HABM to determine the local hypothesis parameters. Such ``unused" measurement may correspond to either the first-time detection of a newborn object or a false detection. To handle false detections, HABM filters out unused measurements with detection scores $z_s$ below a predefined threshold $\eta_\mathrm{score}$ and collect them as $\{\mathbf{z}_k^m\}_{m\in\mathbf{M}_k^\mathrm{low}}$, where $\mathbf{M}_k^\mathrm{low}$ represents the index set for low-score measurements. During the PMB update step, a low-score measurement $\mathbf{z}_k^m$ is assumed to originate from clutter, and the first-time detection hypothesis is parameterized as
\begin{equation}
\label{Clutter Hypothesis}
\begin{cases}
    w=-\ln[\lambda^\mathrm{c}(\mathbf{z}_k^m)]\\
    r= 0.
\end{cases}
\end{equation}
Note the spatial distribution $p(\mathbf{x})$ is not specified, as this local hypothesis is directly removed during the pruning step and does not contribute to any object track. 

For the remaining unused measurements with detection scores greater than or equal to the threshold $\eta_\mathrm{score}$, i.e., $\{\mathbf{z}_k^m\}_{m\in\mathbf{M}_k^\mathrm{high}}$, HABM treats them as first-time detections of newborn objects. Each high-score measurement is associated with a specialized PPP intensity defined as
\begin{equation}
    \lambda^\mathrm{b0}(\mathbf{x}) = \mu^\mathrm{b0}(c) p_\mathrm{uni}(\mathbf{x}).
\end{equation}
where $p_\mathrm{uni}(\mathbf{x})$ represents a uniform spatial distribution over the region of observation, $\mu^\mathrm{b0}(c)$ is the predefined birth rate parameter for object class $c$. This special PPP intensity $\lambda^\mathrm{b0}(\mathbf{x})$ represents undetected objects within the observation region at the previous time step. It is assumed to be time-invariant and provides no prior information about the object's motion state. Therefore, $ \lambda^\mathrm{b0}(\mathbf{x})$ functions as an independent density in HABM and is excluded from the PMB density recursions. The first-time detection hypothesis generated for a high-score measurement $\mathbf{z}_k^m$ is determined by
\begin{equation}
\label{Newborn Hypothesis}
\begin{cases}
    w=-\ln[\mu^\mathrm{b0}(z_{c,k}^m)\frac{1-p_\mathrm{a}(\mathbf{z}_k^m)}{A} + \lambda^\mathrm{c}(\mathbf{z}_k^m)]\\
    r=1\\
    p(\mathbf{x})=\mathcal{N}(\mathbf{x};\mathbf{m}_k^{0,m},P_k^{0,m})
\end{cases}
\end{equation}
where $A$ denotes the area size of the observation region, $p_\mathrm{a}(\mathbf{z}_k^m)$ is the estimated association probability between $\mathbf{z}_k^m$ and existing objects, defined as:
\begin{equation}
\label{Association Probability}
\begin{aligned}
    p_\mathrm{a}(\mathbf{z}_k^m) &= \min[1,\textstyle\sum_{n=1}^{N^\mathrm{d}_{k-1}} \max(\{\mathcal{L}^{m,i}_k\}_{i\in \mathbf{L}_{k-1}^n})]\\
    \mathcal{L}^{m,i}_k &= \mathcal{N}(\mathbf{z}_{xy,k}^m;\hat{\mathbf{z}}^i_{k},S^i_k).
\end{aligned}
\end{equation}
The calculation method for the likelihood $\mathcal{L}^{m,i}_k$ is provided in \eqref{Detection Hypothesis Update} and \eqref{Predicted Measurement} in Section \ref{Misdetect and Detection Hypotheses}. According to the definition of \eqref{Newborn Hypothesis}, if the unused measurement $\mathbf{z}_k^m$ has a higher likelihood of being associated with existing objects, the first-time detection hypothesis will incur a higher association cost. The Gaussian spatial distribution $p(\mathbf{x})$ in \eqref{Newborn Hypothesis} is characterized by the mean vector $\mathbf{m}_k^{0,m}$ and the covariance matrix $P_k^{0,m}$ predefined for each object class
\begin{equation}
\begin{cases}
    \mathbf{m}_k^{0,m}=[z_{x,k}^m,z_{y,k}^m,\sqrt{(z_{vx,k}^m)^2 + (z_{vy,k}^m)^2}, \\
    \qquad \qquad \qquad \qquad \arctan(z_{vy,k}^m,z_{vx,k}^m), 0, 0]^\mathrm{T}\\
    P_k^{0,m}=P^0(z_{c,k}^m).
\end{cases}
\end{equation}
This adaptive design helps reduce false track initializations while allowing OptiPMB to promptly initialize new tracks when measurements with high detection scores are observed.

To enhance recall performance and reduce false negative errors, HABM creates a Poisson component for each low-score measurement and defines the PPP intensity for newborn undetected potential objects as
\begin{equation}
\label{HABM}
\begin{aligned}
    \lambda_k^\mathrm{b}(\mathbf{x}) &= \textstyle\sum_{m\in \mathbf{M}_k^\mathrm{low}} \mu_k^{\mathrm{b},m} \mathcal{N}(\mathbf{x};\mathbf{m}_{k}^{\mathrm{0},m},P_{k}^{\mathrm{0},m})\\
    \mu_k^{\mathrm{b},m} &= \mu^\mathrm{ab}(z_{c,k}^m)[1-p_\mathrm{a}(\mathbf{z}_k^m)]
\end{aligned}
\end{equation}
where $\mathbf{M}_k^\mathrm{low}$ is the index set for low-score measurements, and $\mu^\mathrm{ab}(z_{c,k}^m)$ denotes the adaptive birth rate parameter predefined for each object class. If a low-score measurement originates from a newborn object rather than clutter, its corresponding Poisson component in $\lambda_k^\mathrm{b}(\mathbf{x})$ is likely to be associated with subsequent measurements and initialize a new track later. This measurement-driven definition of $\lambda_k^\mathrm{b}(\mathbf{x})$ allows OptiPMB to effectively account for measurements with low detection scores while reducing the risk of missing newborn objects.

\subsection{Optimizations}
\label{Optimizations}
To further enhance 3D MOT performance, we optimize the conventional PMB filter-based tracking pipeline by redesigning two key algorithm modules in OptiPMB.

\subsubsection{\textbf{Redundant PPP Pruning (RPP)}}
\label{Redundant PPP Pruning}
Although the HABM proposed in OptiPMB can reduce false negative errors by introducing the measurement-driven newborn PPP intensity $\lambda_k^\mathrm{b}(\mathbf{x})$, the number of Poisson components within $\lambda_k^\mathrm{u}(\mathbf{x})$ increases over time. To improve computational efficiency and reduce false track initialization, removing redundant Poisson components during the PPP pruning step is necessary. While the previous PMBM-based 3D MOT methods \cite{IV2018PMBM, RFS_M3} did not specify a dedicated pruning strategy, the original PMBM filter \cite{point_target_MOT_PMBM} only prunes the Poisson components with weights below a predefined threshold. However, this approach is not well-suited for measurement-driven PPP birth modeling. Since the PPP removal solely depends on one pruning threshold, new Poisson components with low initial weights may be pruned before they can correctly initialize a new track, while Poisson components that have already generated detected potential objects may persist unnecessarily, leading to false track initiations. The GNN-PMB tracker \cite{GNN_PMB} utilizes a full measurement-driven approach for generating $\lambda_k^\mathrm{b}(\mathbf{x})$ but removes all Poisson components at the pruning step. This aggressive pruning strategy may result in track initialization failures. 

OptiPMB proposes a new pruning strategy to address these limitations. Specifically, the PPP pruning module marks Poisson components that generate any first-time detection hypothesis during the PMB update step and assigns a counter variable to each Poisson component to track its survival duration. After the PPP prediction and update steps, the pruning module removes all marked Poisson components to prevent repeated track initialization. Additionally, any remaining Poisson components that have persisted for more than $\eta_\mathrm{step}$ time steps are discarded. This PPP pruning mechanism balances between reliable track initialization and computational efficiency, benefiting MOT performance in complex scenarios.

\subsubsection{\textbf{Optimized Track Extraction (OTE)}}
\label{Track Extraction}
After performing PMB prediction and update, data association, and PMB pruning steps, OptiPMB extracts object states from the posterior PMB density $f(\mathbf{X}_k|\mathbf{Z}_{1:k})$ and outputs the estimated object tracks at current time. The previous PMBM/PMB-based trackers \cite{IV2018PMBM, RFS_M3, GNN_PMB} apply an existing probability filter to extract tracks, where the Bernoulli component $\mathbf{X}^i_k$ with an existing probability $r^i_k$ exceeding a predefined threshold $\eta_\mathrm{ext}$ are selected as object tracks. However, relying on a single extraction threshold makes it difficult to achieve both fast track initialization and efficient track termination simultaneously, as a high threshold may delay track initialization, while a low threshold may cause misdetected or lost tracks persist longer than necessary and result in false positive errors.

To improve the flexibility of track extraction, OptiPMB introduces a misdetection counter and redesigns the existing probability filter by applying two extraction thresholds $\eta_\mathrm{ext1}$ and $\eta_\mathrm{ext2}$, satisfying $\eta_\mathrm{ext1}\leq \eta_\mathrm{ext2}$. The IDs of extracted tracks are recorded in a set $\mathbf{T}_k$. Consider a Bernoulli component $\mathbf{X}^i_k$, the track extraction logic is as follows:
\begin{itemize}
    \item If the track corresponding to $\mathbf{X}^i_k$ has not been extracted previously, i.e., $t^i_k\notin \mathbf{T}_{k-1}$, then $\mathbf{X}^i_k$ is extracted as an object track if its existence probability satisfies $r^i_k\geq \eta_\mathrm{ext1}$.
    \item If $t^i_k\in \mathbf{T}_{k-1}$, $\mathbf{X}^i_k$ is extracted only if $r^i_k\geq \eta_\mathrm{ext2}$ and the misdetection counter variable $x_{\mathrm{cnt},k}^i$ is smaller than a predefined limit $\eta_\mathrm{cnt}$.
\end{itemize}
This redesign of track extraction strategy allows faster track initialization with a lower extraction threshold $\eta_\mathrm{ext1}$. Additionally, when an existing object is misdetected over multiple time steps, the introduction of $\eta_\mathrm{ext2}$ and $\eta_\mathrm{cnt}$ enables faster track termination, thereby reducing false positive errors.

\section{Experiments and Performance Analysis}
\label{Evaluations}
\subsection{Dataset and Implementation Details}
The proposed OptiPMB tracker is compared with published and peer-reviewed state-of-the-art online 3D MOT methods on two widely used open-source datasets: the nuScenes dataset \cite{nuScenes} and the KITTI dataset \cite{KITTI}. The official nuScenes tracking task evaluates 3D MOT performance across seven object categories (bicycle, bus, car, motorcycle, pedestrian, trailer, and truck) using average multi-object tracking accuracy (AMOTA) and average multi-object tracking precision (AMOTP) \cite{AB3DMOT} as primary evaluation metrics. The official KITTI object tracking benchmark evaluates 2D MOT performance, which involves tracking 2D object bounding boxes in camera coordinates for car and pedestrian categories. It primarily uses high-order tracking accuracy (HOTA) \cite{HOTA} as the main evaluation metric. 
To provide a more comprehensive performance analysis, we extend the evaluation metrics as follows:
\begin{itemize}
    \item \textbf{nuScenes Dataset}: We incorporate the HOTA metric to further evaluate 3D MOT performance. Specifically, HOTA evaluates tracking accuracy by matching tracking results and ground truths based on a similarity score $\mathcal{S}$ \cite{HOTA}. To adapt HOTA for 3D MOT evaluation, we define the similarity score $\mathcal{S}$ as:
\begin{equation}
    \mathcal{S}=\max[0,\ 1-D(\hat{\mathbf{x}}_{xy},\mathbf{x}_{xy})/D_0]
\end{equation}
where $D(\hat{\mathbf{x}}_{xy},\mathbf{x}_{xy})$ calculates the Euclidean distance between an object's estimated 2D center position $\hat{\mathbf{x}}_{xy}$ and the ground truth center position $\mathbf{x}_{xy}$ on the ground plane, $D_0$ is the distance threshold, beyond which the similarity score reduces to zero. For the nuScenes dataset evaluation, we set the distance threshold as $D_0=2\textrm{m}$, meaning that an estimated object can be matched with a ground truth if their 2D center distance satisfies $D(\hat{\mathbf{x}}_{xy},\mathbf{x}_{xy})\leq2\text{m}$. This matching criterion is consistent with the official nuScenes tracking task settings\footnote{\href{https://www.nuscenes.org/tracking}{https://www.nuscenes.org/tracking}}, ensuring that our HOTA-based evaluation aligns with standard benchmarks for fair comparison.
    \item \textbf{KITTI Dataset}: We evaluate 3D MOT accuracy using the multi-object tracking accuracy (MOTA), AMOTA, and scaled AMOTA (sAMOTA) metrics, following the widely accepted protocol proposed in \cite{AB3DMOT}.
\end{itemize}
Other commonly used secondary metrics, including detection accuracy score (DetA), association accuracy score (AssA), multi-object tracking precision (MOTP), true positives (TP), false positives (FP), false negatives (FN), and ID switches (IDS), 
are also reported in the following evaluations (see \cite{nuScenes,AB3DMOT,HOTA} for definitions of these metrics). As shown in Table \ref{parameters}, the parameters of OptiPMB are finetuned on the nuScenes and KITTI training sets using CenterPoint \cite{CenterPoint} and CasA \cite{CasA} detections, respectively.

\begin{figure*}[t]
\centering
\includegraphics[width=1\textwidth]{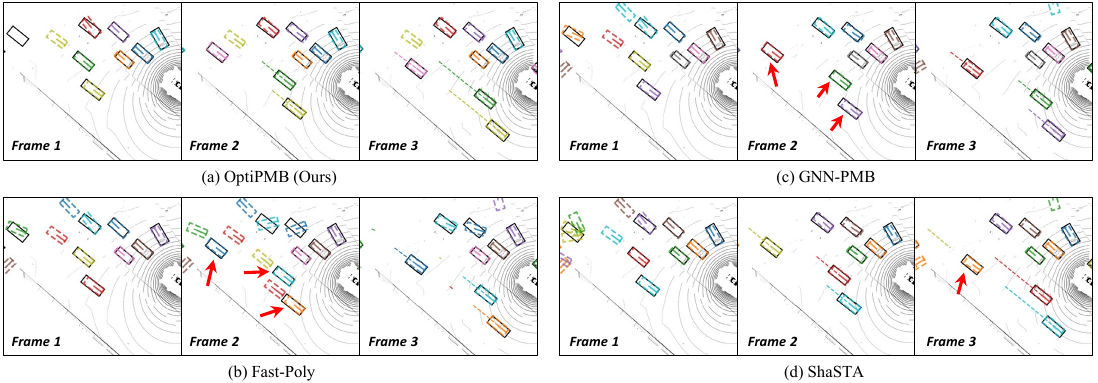}
\caption{Qualitative comparison among our proposed OptiPMB, Fast-Poly \cite{Fast_Poly}, GNN-PMB \cite{GNN_PMB}, and ShaSTA \cite{ShaSTA} on the nuScenes validation set (Scene-0095). Black solid boxes denote ground truth objects. Dashed boxes represent estimated objects, with distinct colors indicating different IDs. Dashed colored lines depict object trajectories. ID switch errors are highlighted by \textcolor{red}{red} arrows.}
\label{Fig4}
\vspace{-0.2cm}
\end{figure*}

\begin{table}[t]
    \belowrulesep=0pt
    \aboverulesep=0pt
    \centering
    \footnotesize
    \caption{Finetuned OptiPMB parameters.}
    \label{parameters}
    \renewcommand\tabcolsep{0.5pt}
\begin{threeparttable}
    \begin{tabular}{c| ccccccc | c}
    \toprule
    \multirow{2}{*}{\textbf{Parameters}} & 
    \multicolumn{7}{c|}{\textbf{nuScenes Category}} & \textbf{KITTI} \\
    \cline{2-9} 
    & Bic. & Bus & Car & Mot. & Ped. & Tra. & Tru. &  Car \\
    \hline
    Score filter threshold $\eta_\mathrm{sf}$           & 0.15    & 0   & 0.1   & 0.16  & 0.2   & 0.1   & 0     & 0.3 \\
    NMS IoU threshold $\eta_\mathrm{iou}$               & 0.1     & 0.1 & 0.1   & 0.1   & 0.1   & 0.1   & 0.1   & 0.1 \\
    Survival probability $p_{\mathrm{s}}$               & 0.99    & 0.99& 0.99  & 0.99  & 0.99  & 0.99  & 0.99  & 1 \\
    Gating distance threshold $\eta_\mathrm{dist}$      & 3       & 10  & 10    & 4     & 3     & 10    & 10    & 4 \\
    Baseline detection probability $p_{\mathrm{d}0}$    & 0.8     & 0.9 & 0.9   & 0.8   & 0.8   & 0.9   & 0.9   & 0.9 \\
    Expected number of points $\mathrm{PTS}_0$          & 10      & 10  & 10    & 10    & 10    & 10    & 10    & 10 \\
    Minimal scaling factor $s_\mathrm{d}$               & 0.5     & 0.5 & 0.5   & 0.5   & 0.5   & 0.5   & 0.5   & 0.7 \\
    HABM score threshold $\eta_\mathrm{score}$          & 0.17    & 0.3 & 0.25  & 0.18  & 0.2   & 0.15  & 0.15  & 0.65 \\
    Adaptive birth rate $\mu^\mathrm{ab}$               & 2       & 2   & 2     & 2     & 2     & 2     & 2     & 0.2 \\
    Undetected object birth rate $\mu^\mathrm{b0}$      & 1       & 5   & 2     & 1     & 1     & 2     & 2     & 0.5 \\
    Clutter rate $\mu^\mathrm{c}$                       & 0.5     & 0.2 & 1     & 0.5   & 0.5   & 0.5   & 1     & 10 \\
    PPP max survival time steps $\eta_\mathrm{step}$    & 3       & 3   & 3     & 2     & 2     & 2     & 2     & 1 \\
    Track extraction threshold 1 $\eta_\mathrm{ext1}$   & 0.7     & 0.7 & 0.7   & 0.7   & 0.7   & 0.7   & 0.5   & 0.85 \\
    Track extraction threshold 2 $\eta_\mathrm{ext2}$   & 0.95    & 0.7 & 0.8   & 0.95  & 0.8   & 0.8   & 0.9   & 0.9 \\
    Misdetection counter limit $\eta_\mathrm{cnt}$      & 3       & 2   & 2     & 2     & 2     & 2     & 2     & 5 \\
    \bottomrule
    \end{tabular}
\end{threeparttable}
\end{table}

\subsection{Comparison with State-of-the-Arts}
\subsubsection{nuScenes Dataset}
As shown in Table \ref{nuScenes test performance}, our proposed OptiPMB demonstrates state-of-the-art online 3D MOT performance on the nuScenes test set. Specifically, it achieves overall AMOTA scores of 0.767 using the LargeKernel3D \cite{LargeKernel3D} detector and 0.713 using the CenterPoint \cite{CenterPoint} detector. With the same detectors, OptiPMB outperforms the previous learning-assisted \cite{OGR3MOT,NEBP}, learning-based \cite{3DMOTFormer,ShaSTA} and model-based \cite{RFS_M3,SimpleTrack,GNN_PMB,Poly_MOT,Fast_Poly} methods in terms of the primary tracking accuracy metric, AMOTA. Our method also demonstrates strong performance in secondary metrics, including TP, FN, FP, and IDS, especially comparing with the previous model-based RFS trackers, the RFS-M$^3$ \cite{RFS_M3} and GNN-PMB \cite{GNN_PMB}. Our method is fully open-sourced, with code and parameters available\footnote{\href{https://github.com/dinggh0817/OptiPMB}{Code is available at: https://github.com/dinggh0817/OptiPMB}}, to demonstrate the potential of RFS-based trackers and serve as a new baseline for model-based 3D MOT.

To comprehensively evaluate our proposed method, we further compare OptiPMB with other state-of-the-art 3D MOT methods across all seven object categories on the nuScenes validation set with extended metrics. For a fair comparison, we select three open-source trackers as baselines: Fast-Poly \cite{Fast_Poly}, GNN-PMB \cite{GNN_PMB}, and ShaSTA \cite{ShaSTA}. These trackers all employ the TBD strategy and provide parameters finetuned for the CenterPoint detector while representing different design approaches of 3D MOT methods:
\begin{itemize}
    \item \textbf{Fast-Poly}: A state-of-the-art model-based tracker using two-stage data association and confidence score-based track management. Its design can be categorized as TBD with RV-based Bayesian filter (Section \ref{TBD with RV Bayesian filter}).
    \item \textbf{GNN-PMB}: The previous state-of-the-art PMB-based tracker, categorized as TBD with RFS-based Bayesian filter (Section \ref{TBD with RFS Bayesian filter}).
    \item \textbf{ShaSTA}: A learning-assisted TBD tracker using spatial-temporal and shape affinities to improve association quality. Its design can be categorized as TBD with learning-assisted data association (Section \ref{Learning Embedding for TBD with Bayesian filter}).
\end{itemize}

% \FloatBarrier

\begin{table*}[t]
\belowrulesep=0pt
\aboverulesep=0pt
\footnotesize
\caption{Online 3D MOT performance of OptiPMB and other advanced trackers on the nuScenes test set. Trackers using CenterPoint \cite{CenterPoint} and LargeKernel3D \cite{LargeKernel3D} detectors are distinguished for fair comparison, with different input modalities (C: camera, L: LiDAR) indicated. \textbf{Bold} and \underline{underline} denote the first and second best results among trackers using the same detector. \textcolor[RGB]{0, 204, 204}{Blue} highlights the primary tracking accuracy metric, AMOTA.
}
\label{nuScenes test performance}
\renewcommand\tabcolsep{1.5pt}
\begin{center}
\begin{threeparttable}
\begin{tabular}{c|c|c| >{\columncolor[RGB]{179, 255, 255}}cccccccc | c c c c c}
\toprule
\multirow{2}{*}{\textbf{Method}} & \multirow{2}{*}{\textbf{Detector}} & \multirow{2}{*}{\textbf{Modal.}} & \multicolumn{8}{c|}{\textbf{AMOTA$\uparrow$}} & \multirow{2}{*}{\textbf{AMOTP$\downarrow$}} & \multirow{2}{*}{\textbf{TP$\uparrow$}} & \multirow{2}{*}{\textbf{FN$\downarrow$}} & \multirow{2}{*}{\textbf{FP$\downarrow$}} & \multirow{2}{*}{\textbf{IDS$\downarrow$}} \\
\cline{4-11}
& & & Overall & Bic. & Bus & Car & Mot. & Ped. & Tra. & Tru. & & & & & \\
\hline

\makecell[c]{RFS-M$^3$ (ICRA 2021) \cite{RFS_M3}}
& CenterPoint & L       & 0.619 & N/A & N/A & N/A & N/A & N/A & N/A & N/A & 0.752 & 90872 & 27168 & \underline{16728} & 1525 \\
\makecell[c]{OGR3MOT (RA-L 2022) \cite{OGR3MOT}} & CenterPoint & L       & 0.656 & 0.380 & 0.711 & 0.816 & 0.640 & 0.787 & 0.671 & 0.590 & 0.620 & 95264 & 24013 & 17877 & 288 \\
\makecell[c]{SimpleTrack (ECCVW 2022) \cite{SimpleTrack}}
& CenterPoint & L       & 0.668 & 0.407 & 0.715 & 0.823 & 0.674 & 0.796 & 0.673 & 0.587 & 0.550 & 95539 & 23451 & 17514 & 575 \\
\makecell[c]{GNN-PMB (T-IV 2023) \cite{GNN_PMB}}
& CenterPoint & L       & 0.678 & 0.337 & 0.744 & \underline{0.839} & 0.657 & 0.804 & \underline{0.715} & 0.647 & 0.560 & 97274 & 21521 & 17071 & 770 \\
\makecell[c]{3DMOTFormer (ICCV 2023) \cite{3DMOTFormer}}
& CenterPoint & L       & 0.682 & 0.374 & \textbf{0.749} & 0.821 & 0.705 & 0.807 & 0.696 & 0.626 & \textbf{0.496} & 95790 & 23337 & 18322 & 438 \\
\makecell[c]{NEBP (T-SP 2023) \cite{NEBP}}
& CenterPoint & L       & 0.683 & \textbf{0.447} & 0.708 & 0.835 & 0.698 & 0.802 & 0.690 & 0.598 & 0.624 & 97367
 & 21971 & 16773 & \textbf{227} \\
\makecell[c]{ShaSTA (RA-L 2024) \cite{ShaSTA}}
& CenterPoint & L       & \underline{0.696} & 0.410 & 0.733 & 0.838 & \underline{0.727} & \underline{0.810} & 0.704 & \underline{0.650} & \underline{0.540} & \underline{97799} & \underline{21293} & 16746 & 473 \\
\makecell[c]{OptiPMB (Ours)}
& CenterPoint & L       & \textbf{0.713} & \underline{0.414} & \underline{0.748} & \textbf{0.860} & \textbf{0.751} & \textbf{0.836} & \textbf{0.727} & \textbf{0.658} & 0.541 & \textbf{98550} & \textbf{20758} & \textbf{16566} & \underline{257} \\

\hline
\makecell[c]{Poly-MOT (IROS 2023) \cite{Poly_MOT}}
& LargeKernel3D & C+L   & 0.754 & \underline{0.582} & \textbf{0.786} & \underline{0.865} & 0.810 & 0.820 & 0.751 & \underline{0.662} & \textbf{0.422} & \underline{101317} & \underline{17956} & 19673 & \underline{292} \\
\makecell[c]{Fast-Poly (RA-L 2024) \cite{Fast_Poly}}
& LargeKernel3D & C+L   & \underline{0.758} & 0.573 & 0.767 & 0.863 & \underline{0.826} & \textbf{0.852} & \underline{0.768} & 0.656 & 0.479 & 100824 & 18415 & \textbf{17098} & 326 \\
\makecell[c]{OptiPMB$^\text{\textdagger}$ (Ours)} % Without MCTrack
& LargeKernel3D & C+L   & \textbf{0.767} & \textbf{0.600} & \underline{0.774} & \textbf{0.870} & \textbf{0.827} & \underline{0.848} & \textbf{0.775} & \textbf{0.676} & \underline{0.460} & \textbf{102899} & \textbf{16411} & \underline{18796} & \textbf{255} \\

\bottomrule
\end{tabular}
\begin{tablenotes}
        \footnotesize
        \item[$\text{\textdagger}$] OptiPMB ranked first by overall AMOTA on the \href {https://eval.ai/web/challenges/challenge-page/476/leaderboard/1321}{\textit{nuScenes tracking challenge leaderboard}} at the submission time of our manuscript (Mar. 2025).
\end{tablenotes}
\end{threeparttable}
\end{center}
\vspace{-0.2cm}
\end{table*}

\begin{table*}[t!]
\belowrulesep=0pt
\aboverulesep=0pt
\footnotesize
\caption{Online 3D MOT performance of OptiPMB and other state-of-the-art trackers on the nuScenes validation set. The CenterPoint \cite{CenterPoint} detector is employed for all trackers for a fair comparison. \textbf{Bold} and \underline{underline} denote the first and second best results within each category. \textcolor[RGB]{0, 204, 204}{Blue} indicates the two primary tracking accuracy metrics, HOTA and AMOTA.
}
\label{nuScenes val performance}
\renewcommand\tabcolsep{4pt}
\begin{center}
\begin{threeparttable}
\begin{tabular}{ c|c|>{\columncolor[RGB]{179, 255, 255}}ccc|>{\columncolor[RGB]{179, 255, 255}}cccccc}
  \toprule
  \textbf{Method} & \textbf{Category} & \textbf{HOTA}(\%)$\uparrow$ & \textbf{DetA}(\%)$\uparrow$ & \textbf{AssA}(\%)$\uparrow$ &  \textbf{AMOTA}$\uparrow$ & \textbf{AMOTP}$\downarrow$ & \textbf{TP}$\uparrow$ & \textbf{FN}$\downarrow$ & \textbf{FP}$\downarrow$ & \textbf{IDS}$\downarrow$\\
\hline
\multirow{8}{*}{\makecell[l]{GNN-PMB$^\text{\textdagger}$ (IEEE T-IV 2023) \cite{GNN_PMB}}}
& Bic.      & \textbf{17.88} & \textbf{4.82} & 66.54 & 0.513 & 0.602 & 1135 & 855 & 220 & 3 \\
& Bus       & \underline{46.83} & \underline{30.16} & 72.97 & 0.854 & 0.546 & \textbf{1789} & \textbf{312} & 155 & 11 \\
& Car       & 57.53 & 43.22 & 77.37 & 0.849 & 0.387 & 49182 & 8791 & \underline{6140} & 344 \\
& Mot.      & 21.38 & 6.91 & 66.22 & 0.723 & 0.475 & 1494 & 477 & 247 & 6 \\
& Ped.      & 46.49 & 29.86 & 72.48 & 0.807 & \underline{0.357} & 21149 & 4008 & 3669 & 266 \\
& Tra.      & 23.90 & 9.03 & 64.22 & 0.506 & 0.973 & 1378 & 1045 & 412 & 2 \\ 
& Tru.      & 36.13 & \underline{17.38} & 75.36 & 0.695 & 0.581 & 7007 & 2625 & \textbf{1519} & 18 \\  \hhline{|~|-|-|-|-|-|-|-|-|-|-|}
& Overall   & 46.03 & 28.15 & 75.45 & 0.707 & 0.560 & 83134 & 18113 & \underline{12362} & 650 \\ \hline
\multirow{8}{*}{\makecell[l]{ShaSTA$^\text{\textdagger}$ (IEEE RA-L 2024) \cite{ShaSTA}}} 
& Bic.      & 13.62 & 2.58 & 72.12 & \textbf{0.588} & 0.507 & 1187 & 804 & \textbf{143} & 2 \\
& Bus       & 45.17 & 26.85 & 76.16 & \underline{0.856} & \underline{0.527} & 1777 & 333 & \underline{142} & \underline{2} \\
& Car       & 57.05 & 41.48 & 78.67 & 0.856 & \textbf{0.337} & 49200 & 8986 & \textbf{6067} & \underline{131} \\
& Mot.      & 21.56 & 6.83 & 68.10 & 0.745 & 0.521 & 1457 & 513 & \textbf{83} & 7 \\
& Ped.      & 46.99 & 29.48 & 74.95 & 0.814 & 0.378 & 20945 & 4258 & \textbf{3378} & 220 \\
& Tra.      & 23.69 & 8.63 & 65.84 & \underline{0.531} & 0.970 & 1284 & 1139 & \textbf{238} & 2 \\ 
& Tru.      & \underline{36.24} & 17.17 & 76.70 & 0.703 & 0.572 & \textbf{7222} & \textbf{2421} & 1671 & 7 \\ \hhline{|~|-|-|-|-|-|-|-|-|-|-|}
& Overall   & 43.87 & 24.98 & 77.18 & 0.728 & 0.544 & 83072 & 18454 & \textbf{11722} & \underline{371} \\ \hline
\multirow{8}{*}{\makecell[l]{Fast-Poly$^\text{\textdagger}$ (IEEE RA-L 2024) \cite{Fast_Poly}}} 
& Bic.      & 16.36 & 3.61 & \textbf{74.30} & 0.572 & 0.553 & 1218 & 775 & \underline{189} & \textbf{0} \\
& Bus       & 46.30 & 27.97 & \underline{76.90} & 0.855 & \underline{0.522} & \underline{1787} & \underline{316} & 172 & 9 \\
& Car       & \underline{62.63} & \textbf{49.48} & \underline{79.56} & \underline{0.860} & \underline{0.339} & \textbf{50443} & \textbf{7654} & 8057 & 220 \\
& Mot.      & \underline{27.44} & \underline{9.95} & \textbf{75.83} & \textbf{0.792} & \textbf{0.470} & \textbf{1559} & \textbf{415} & 178 & \textbf{3} \\
& Ped.      & \underline{57.59} & \underline{43.05} & \underline{77.12} & \underline{0.835} & \textbf{0.355} & \textbf{21916} & \textbf{3332} & 4070 & \underline{175} \\
& Tra.      & \underline{24.69} & \underline{9.29} & \underline{66.40} & 0.529 & \textbf{0.939} & \underline{1440} & \underline{984} & \underline{371} & \underline{1} \\ 
& Tru.      & 32.29 & 13.40 & \underline{78.00} & \underline{0.714} & \textbf{0.533} & \underline{7220} & \underline{2424} & 1676 & \underline{6} \\ \hhline{|~|-|-|-|-|-|-|-|-|-|-|}
& Overall   & \underline{48.52} & \underline{30.04} & \underline{78.55} & \underline{0.737} & \textbf{0.530} & \textbf{85583} & \textbf{15900} & 14713 & 414 \\ \hline
\multirow{8}{*}{\makecell[l]{OptiPMB (Ours)}} 
& Bic.      & \underline{17.50} & \underline{4.20} & \underline{73.07} & \underline{0.575} & \underline{0.514} & \textbf{1221} & \textbf{771} & 208 & \underline{1} \\
& Bus       & \textbf{55.91} & \textbf{40.69} & \textbf{77.27} & \textbf{0.857} & 0.538 & 1785 & 327 & \textbf{131} & \textbf{0} \\
& Car       & \textbf{62.69} & \underline{49.10} & \textbf{80.47} & \textbf{0.870} & 0.359 & \underline{50396} & \underline{7864} & 6850 & \textbf{57} \\
& Mot.      & \textbf{32.52} & 14.06 & \underline{75.39} & \textbf{0.792} & \textbf{0.470} & \underline{1497} & \underline{477} & \underline{112} & \textbf{3} \\
& Ped.      & \textbf{61.73} & \textbf{49.41} & \textbf{77.24} & \textbf{0.838} & \underline{0.357} & \underline{21365} & \underline{3892} & \underline{3401} & \textbf{166} \\
& Tra.      & \textbf{27.12} & \textbf{10.59} & \textbf{70.70} & \textbf{0.534} & \underline{0.949} & \textbf{1488} & \textbf{937} & 399 & \textbf{0} \\ 
& Tru.      & \textbf{38.77} & \textbf{19.05} & \textbf{79.21} & \textbf{0.718} & \underline{0.539} & 7211 & 2439 & \underline{1576} & \textbf{1} \\ \hhline{|~|-|-|-|-|-|-|-|-|-|-|}
& Overall   & \textbf{52.06} & \textbf{34.31} & \textbf{79.27} & \textbf{0.741} & \underline{0.532} & \underline{84963} & \underline{16707} & 12677 & \textbf{227} \\ 
\bottomrule
\end{tabular}

\begin{tablenotes}
    \footnotesize
    \item[$\text{\textdagger}$] Evaluated using the authors' code and parameters. The tracking performance is identical to that reported in the original manuscripts.
\end{tablenotes}

\end{threeparttable}
\end{center}
\vspace{-0.2cm}
\end{table*}

As demonstrated in Table \ref{nuScenes val performance}, with the same CenterPoint detector, OptiPMB achieves the highest overall HOTA (52.06\%) and AMOTA (0.741) among the compared methods, indicating its superior tracking accuracy. OptiPMB surpasses other methods in HOTA, AMOTA, and IDS across all object categories, except for the bicycle category, where it ranks second. Notably, compared to GNN-PMB, OptiPMB significantly improves tracking accuracy (+6.03\% HOTA and +3.4\% AMOTA) and track ID maintenance (-423 IDS) while utilizing a similar PMB filter-based RFS framework. This performance difference underscores the effectiveness of the innovative designs and modules proposed in OptiPMB. 
HOTA and its sub-metrics (AssA and DetA) in Table \ref{nuScenes val performance} are evaluated across multiple localization accuracy thresholds without considering object confidence scores or applying track interpolation post-processing \cite{HOTA}. Consequently, the DetA and AssA scores indicate that OptiPMB achieves high detection and association accuracy while evaluating all raw tracking results. A qualitative comparison in Fig. \ref{Fig4} also illustrates the superior tracking performance of OptiPMB. In a challenging scenario where high-velocity cars enter the observation area, OptiPMB correctly initiates and maintains tracks. In contrast, the model-based methods Fast-Poly and GNN-PMB fail to associate the new objects with subsequent detections, leading to ID switch errors. The learning-assisted tracker ShaSTA exhibits fewer ID switches but repeatedly initializes tracks in the first frame.

\FloatBarrier

\subsubsection{KITTI Dataset} Since the official KITTI object tracking evaluation only includes 2D MOT metrics, we compare the online 3D MOT performance of OptiPMB and other advanced trackers on the car category of the KITTI validation dataset, following the evaluation protocol proposed in \cite{AB3DMOT}. This protocol designates sequences 1, 6, 8, 10, 12, 13, 14, 15, 16, 18, and 19 in the original KITTI training set as the validation set. As shown in Table \ref{KITTI validation performance}, OptiPMB achieves superior tracking accuracy on the KITTI validation set, outperforming other state-of-the-art trackers that use the same PointRCNN \cite{PointRCNN}, PointGNN \cite{Point-GNN}, and CasA \cite{CasA} detectors in sAMOTA scores. Here, the online 3D MOT results of CasTrack \cite{CasA}, MCTrack \cite{MCTrack}, and RobMOT \cite{RobMOT} are evaluated using their online tracking configurations and finetuned parameters provided by the authors. 
For reference, the 2D MOT performance for the compared trackers on the KITTI test set is presented in Table \ref{KITTI test performance}. With the PointRCNN \cite{PointRCNN} and PointGNN \cite{Point-GNN} detection results, OptiPMB achieves significantly higher tracking accuracy than the baseline LiDAR-only trackers \cite{AB3DMOT, PolarMOT} while generating the fewest false positives and ID switch errors among the compared methods. Although OptiPMB relies solely on LiDAR detection results, its HOTA score is still comparable with camera-LiDAR fusion-based trackers \cite{Feng, DeepFusionMOT, StrongFusionMOT, CAMO_MOT}. Moreover, when utilizing higher-quality CasA \cite{CasA} detection results, OptiPMB surpasses even the recently proposed RobMOT \cite{RobMOT} and MMF-JDT \cite{MMF-JDT} trackers in HOTA, DetA, MOTA, TP, and FP scores, demonstrating a strong tracking performance.

\begin{table*}[t]
\belowrulesep=0pt
\aboverulesep=0pt
%\small
\caption{Online 3D MOT performance of OptiPMB and other advanced trackers on the KITTI car validation set. PointRCNN \cite{PointRCNN}, Point-GNN \cite{Point-GNN}, and CasA \cite{CasA} detectors are distinguished for fair comparison, with different modalities (C for camera, L for LiDAR) indicated. \textbf{Bold} and \underline{underline} denote the first and second best results among trackers using the same detector. 
}
\label{KITTI validation performance}
\renewcommand\tabcolsep{3pt}
\begin{center}
\begin{threeparttable}
\begin{tabular}{ c|c|c|ccccccc}
  \toprule
  \textbf{Method} & \textbf{3D Detector} & \textbf{Modality} & \textbf{sAMOTA}(\%) & \textbf{AMOTA}(\%)$\uparrow$ & \textbf{AMOTP}(\%)$\uparrow$ & \textbf{MOTA}(\%)$\uparrow$ & \textbf{MOTP}(\%)$\uparrow$ & \textbf{IDS}$\downarrow$ \\
\hline
AB3DMOT (IROS 2020) \cite{AB3DMOT} & PointRCNN & L & 93.28 & 45.43 & 77.41 & 86.24 & 78.43 & \textbf{0} \\
ConvUKF (T-IV 2024) \cite{ConvUKF} & PointRCNN & L & 93.32 & 45.46 & \underline{78.09} & N/A & N/A & N/A \\
GNN3DMOT (CVPR 2020) \cite{Gnn3dmot} & PointRCNN & C+L & 93.68 & 45.27 & \textbf{78.10} & 84.70 & \textbf{79.03} & \textbf{0}  \\
FGO-based 3D MOT (S-J 2024) \cite{Real_Time_MOT} & PointRCNN & L & \underline{93.77} & \underline{46.14} & 77.85 & \underline{86.53} & \underline{79.00} & 1 \\
OptiPMB (Ours) & PointRCNN & L & \textbf{93.78} & \textbf{48.40} & 77.30 & \textbf{87.53} & 77.39 & \textbf{0} \\
\hline
PolarMOT (ECCV 2022) \cite{PolarMOT} & Point-GNN & L & 94.32 & N/A & N/A & \textbf{93.93} & N/A & \underline{31} \\
CAMO-MOT (T-ITS 2023) \cite{CAMO_MOT} & Point-GNN & C+L & \underline{95.20} & \underline{48.04} & \textbf{81.48} & N/A & N/A & N/A \\
OptiPMB (Ours) & Point-GNN & L & \textbf{96.29} & \textbf{48.87} & \underline{79.39} & \underline{93.48} & 78.75 & \textbf{0} \\
\hline
RobMOT $^\text{\textdagger}$ (T-ITS 2025) \cite{RobMOT} & CasA & L & 91.16 & 46.95 & \underline{82.06} & 89.53 & 83.57 & \textbf{0} \\
MCTrack$^\text{\textdagger}$ (arXiv 2024) \cite{MCTrack} & CasA & L & 91.69 & \textbf{47.29} & 81.88 & \underline{90.52} & \underline{84.01} & \textbf{0} \\
CasTrack$^\text{\textdagger}$ (T-GRS 2022) \cite{CasA} & CasA & L & \underline{92.14} & 44.94 & \textbf{82.60} & 89.45 & \textbf{84.34} & 2 \\
OptiPMB (Ours) & CasA & L & \textbf{93.40} & \underline{47.19} & 81.47 & \textbf{91.00} & 81.03 & \textbf{0} \\
\bottomrule
\end{tabular}

\begin{tablenotes}
    \footnotesize
    \item[$\text{\textdagger}$] Evaluated using the authors' official code, parameters, and online tracking configurations. 
\end{tablenotes}
\end{threeparttable}
\end{center}
\vspace{-0.2cm}
\end{table*}

\begin{table*}[t]
\belowrulesep=0pt
\aboverulesep=0pt
\caption{Online 2D MOT performance of our proposed OptiPMB and other advanced trackers on the KITTI car test dataset. Trackers using PointRCNN \cite{PointRCNN}, Point-GNN \cite{Point-GNN}, and CasA \cite{CasA} detectors are distinguished for fair comparison, with different modalities (C for camera, L for LiDAR) indicated. \textbf{Bold} and \underline{underline} denote the first and second best results among trackers using the same detector. \textcolor[RGB]{0, 204, 204}{Blue} indicates the primary tracking accuracy metric, HOTA.}
\label{KITTI test performance}
\renewcommand\tabcolsep{2pt}
\begin{center}
\begin{threeparttable}
\begin{tabular}{ c|c|c|>{\columncolor[RGB]{179, 255, 255}}cccccccc}
  \toprule
  \textbf{Method} & \textbf{3D Detector} & \textbf{Modality} & \textbf{HOTA}(\%)$\uparrow$ & \textbf{DetA}(\%)$\uparrow$ & \textbf{AssA}(\%)$\uparrow$ & \textbf{MOTA}(\%)$\uparrow$ & \textbf{MOTP}(\%)$\uparrow$ & \textbf{TP}$\uparrow$ &\textbf{FP}$\downarrow$ & \textbf{IDS}$\downarrow$\\
\hline
AB3DMOT (IROS 2020) \cite{AB3DMOT} & PointRCNN & L & 69.99 & 71.13 & 69.33 & 83.61 & \textbf{85.23} & \underline{29849} & \underline{4543} & 113 \\
Feng et al. (T-IV 2024) \cite{Feng} & PointRCNN & C+L & 74.81 & N/A & N/A & \underline{84.82} & \underline{85.17} & N/A & N/A & N/A \\
DeepFusionMOT (RA-L 2022) \cite{DeepFusionMOT} & PointRCNN & C+L & 75.46 & \underline{71.54} & \underline{80.05} & 84.63 & 85.02 & 29791 & 4601 & 84 \\
StrongFusionMOT (S-J 2023) \cite{StrongFusionMOT} & PointRCNN & C+L & \textbf{75.65} & \textbf{72.08} & 79.84 & \textbf{85.53} & 85.07 & 29734 & 4658 & \underline{58} \\
OptiPMB (Ours) & PointRCNN & L & \underline{75.57} & 70.12 & \textbf{81.90} & 81.97 & 84.73 & \textbf{31763} & \textbf{2629} & \textbf{33} \\
\hline
PolarMOT (ECCV 2022) \cite{PolarMOT} & Point-GNN & L & 75.16 & \underline{73.94} & \underline{76.95} & 85.08 & \textbf{85.63} & \underline{31724} & 2668 & 462 \\
CAMO-MOT (T-ITS 2023) \cite{CAMO_MOT} & Point-GNN & C+L & \textbf{79.95} & N/A & N/A & \textbf{90.38} & \underline{85.00} & N/A & \underline{2322} & \textbf{23}\\
OptiPMB (Ours) & Point-GNN & L & \underline{79.08} & \textbf{75.13} & \textbf{83.86} & \underline{88.76} & 84.90 & \textbf{32554} & \textbf{1838} & \textbf{23} \\
\hline
UG3DMOT (SP 2024) \cite{UG3DMOT} & CasA & L & 78.60 & 76.01 & 82.28 & 87.98 & \textbf{86.56} & 31399 & 2992 & 30 \\
MMF-JDT (RA-L 2025) \cite{MMF-JDT} & CasA & C+L & 79.52 & 75.83 & 84.01 & 88.06 & 86.24 & N/A & N/A & 37 \\
RobMOT (T-ITS 2025) \cite{RobMOT} & CasA & L & \underline{81.22} & \underline{77.48} & \textbf{85.77} & \underline{90.48} & 86.02 & \underline{32670} & \underline{1722} & \textbf{6} \\
OptiPMB (Ours) & CasA & L & \textbf{81.76} & \textbf{78.70} & \underline{85.57} & \textbf{91.57} & \underline{86.47} & \textbf{32909} & \textbf{1483} & \underline{17} \\
\bottomrule
\end{tabular}
\end{threeparttable}
\end{center}
\vspace{-0.2cm}
\end{table*}

\subsection{Ablation Study on OptiPMB}
We conduct ablation experiments on the nuScenes validation set to assess the influence of the proposed components in OptiPMB, as summarized in Table \ref{ablation_performance}. In the baseline tracker, the adaptive detection probability $p_\mathrm{d}(\mathbf{x}_+)$ is replaced with the fixed parameter $p_\mathrm{d0}$, and object tracks with existence probabilities exceeding the threshold $\eta_{\mathrm{ext}1}$ are extracted. In addition, the HABM and OTE modules are replaced by the object birth model and the PPP pruning strategy employed in the GNN-PMB \cite{GNN_PMB} tracker. Except for the detection probability and existence threshold settings described above, all tracker variants share identical parameter configurations. For clarity of comparison, trackers equipped with different numbers of modules are grouped together and ranked within each group according to their AMOTA scores. To analyze the runtime impact of different modules, we also report the average processed frames per second (FPS) of the Python implementation of each tracker variant on our test platform, running Ubuntu 22.04 with an AMD 7950X CPU and 64 GB of RAM. Consistent with common practice in the TBD literature, these FPS values are measured without accounting for detector latency.

As shown in Table \ref{ablation_performance}, HABM is the primary contributor to the effectiveness of OptiPMB, as trackers equipped with HABM achieve the highest AMOTA scores within each group. Incorporating this birth model not only substantially boosts the baseline tracking performance (+3.0\% AMOTA), but also improves both track ID maintenance and runtime efficiency (–36 IDS, +2.6 FPS) by generating more reliable newborn objects and suppressing false potential objects.
However, the other three modules exhibit more complex effects and cannot effectively improve tracking accuracy in isolation. Specifically, adding ADP reduces ID switch errors by enhancing the maintenance of occluded tracks, but it also introduces additional false tracks, which can reduce tracking accuracy. The RPP module reduces false negatives by retaining PPP components for a longer duration, but at the cost of increased tracking latency due to the larger number of potential objects. The OTE module improves accuracy and FPS by reducing false tracks, yet simultaneously increases ID switch errors. Nevertheless, when combined with HABM, the adverse effects of these modules are mitigated, leading to the improvement in overall tracking performance. Consequently, our OptiPMB tracker equipped with all four modules achieves the highest tracking accuracy (+4.5\% AMOTA) among all tracker variants, demonstrating the effectiveness of their joint design.

\begin{table}[t]
    \belowrulesep=0pt
    \aboverulesep=0pt
    \centering
    \footnotesize
    \caption{Ablation study of OptiPMB on the nuScenes validation set using the CenterPoint \cite{CenterPoint} detector.}
    \label{ablation_performance}
    \renewcommand\tabcolsep{2pt}
\begin{threeparttable}
    \begin{tabular}{cccc|ccccccc}
    \toprule
    \multicolumn{4}{c|}{\textbf{Modules}} & 
    \multicolumn{6}{c}{\textbf{Metrics}} \\
    \hhline{|-|-|-|-|-|-|-|-|-|-|-|} 
    ADP & HABM & RPP & OTE
    & AMOTA$\uparrow$ & AMOTP$\downarrow$ & FN$\downarrow$ & FP$\downarrow$ & IDS$\downarrow$  & FPS$\uparrow$ \\
    \midrule
    & & & & 0.696 & 0.622 & 20421 & 12362 & 250 & 17.5\\
    \hline
    \checkmark & & & & 0.688 & 0.609 & 21458 & 13755 & 178 & 15.4\\
    & & \checkmark & & 0.692 & 0.614 & 19664 & 12683 & 325 & 4.7\\
    & & & \checkmark & 0.701 & 0.627 & 20132 & 11020 & 334 & 19.0\\
    & \checkmark & & & 0.726 & 0.551 & 16061 & 14403 & 214 & 20.1\\
    \hline
    \checkmark & & \checkmark & & 0.688 & 0.597 & 20985 & 13820 & 203 & 4.3\\
    & & \checkmark & \checkmark & 0.697 & 0.632 & 19941 & 11082 & 414 & 4.7\\
    \checkmark & & & \checkmark & 0.705 & 0.621 & 20103 & 11301 & 275 & 14.6\\
    \checkmark & \checkmark & & & 0.721 & 0.535 & 17766 & 14975 & 163 & 13.2\\
    & \checkmark & \checkmark & & 0.727 & 0.551 & 17432 & 13097 & 206 & 17.2\\
    & \checkmark & & \checkmark & 0.731 & 0.569 & 17607 & 11385 & 264 & 19.7\\
    \hline
    \checkmark & & \checkmark & \checkmark & 0.708 & 0.615 & 19700 & 11214 & 414 & 4.3\\
    \checkmark & \checkmark & \checkmark & & 0.722 & 0.529 & 17678 & 15046 & 163 & 13.3\\
    & \checkmark & \checkmark & \checkmark & 0.732 & 0.563 & 17595 & 11391 & 264 & 15.6\\
    \checkmark & \checkmark & & \checkmark & 0.739 & 0.534 & 16516 & 12911 & 233 & 13.4\\
    \hline
    \checkmark & \checkmark & \checkmark & \checkmark & 0.741 & 0.532 & 16707 & 12677 & 227 & 11.3\\
    \bottomrule
    \end{tabular}

\end{threeparttable}
% \vspace{-0.3cm}
\end{table}

\begin{figure}[t]
\centerline{\includegraphics[width=0.5\textwidth]{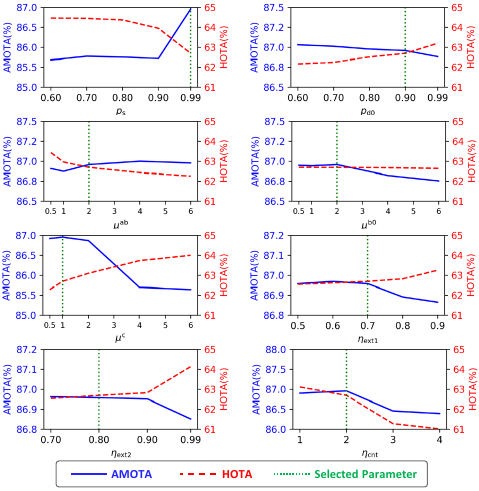}}
\caption{Impact of key OptiPMB parameters on tracking accuracy for cars in the nuScenes validation set with CenterPoint \cite{CenterPoint} detections. Selected parameters are used to report the results in Table \ref{nuScenes val performance}.}
\label{Fig5}
% \vspace{-0.2cm}
\end{figure}

\subsection{Parameter Sensitivity}
The performance of model-based 3D MOT methods often relies on careful selection of hyperparameters. Therefore, we analyzed the impact of key parameters in OptiPMB on the car category in the nuScenes validation set. As shown in Fig. \ref{Fig5}, the survival probability ($p_\mathrm{s}$), clutter rate ($\mu^\mathrm{c}$), and misdetection counter threshold ($\eta_\mathrm{cnt}$) significantly influence both AMOTA and HOTA scores. To achieve better track continuity and recall performance without sacrificing tracking accuracy, we suggest setting these parameters to $p_\mathrm{s}=0.99$, $\mu^\mathrm{c}=1$, and $\eta_\mathrm{cnt}=2$. OptiPMB is less sensitive to other hyperparameters, demonstrating the robustness of our proposed adaptive designs.

\section{Conclusion}
\label{Conclusion}
In this study, we introduced OptiPMB, a novel 3D multi-object tracking (MOT) approach that employs an optimized Poisson multi-Bernoulli (PMB) filter within the tracking-by-detection framework. Our method addressed critical challenges in random finite set (RFS)-based 3D MOT by incorporating a measurement-driven hybrid adaptive birth model, adaptive detection probabilities, and optimized track extraction and pruning strategies. Experimental results on the nuScenes and KITTI datasets demonstrated the effectiveness and robustness of OptiPMB, which achieved state-of-the-art performance compared to existing 3D MOT methods. Consequently, a new benchmark was established for RFS-based 3D MOT in autonomous driving scenarios. Future research will focus on enhancing real-time performance of OptiPMB through parallelization techniques.

%%%%%%%%% REFERENCES

%%% Biography with photos

% \vspace{-0.5cm} 
\begin{IEEEbiography}[{\includegraphics[width=1in,height=1.25in,clip,keepaspectratio]{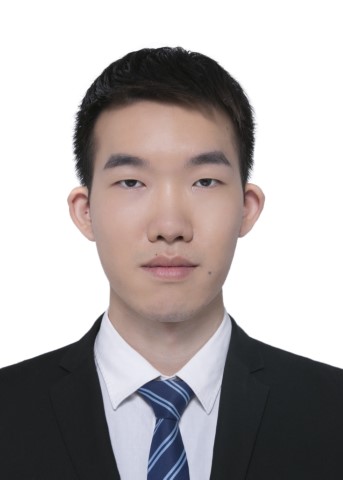}}]{Guanhua Ding}
(Student Member, IEEE) received the B.S. and M.Sc. degrees from Beihang University (BUAA), Beijing, China, in 2020 and 2023, respectively. He is currently pursuing the Ph.D. degree in signal and information processing with the School of Electronic Information Engineering, BUAA, Beijing, China. His research interests include multi-object tracking, extended object tracking, multi-sensor data fusion, and random finite set theory.
\end{IEEEbiography}

\begin{IEEEbiography}[{\includegraphics[width=1in,height=1.25in,clip,keepaspectratio]{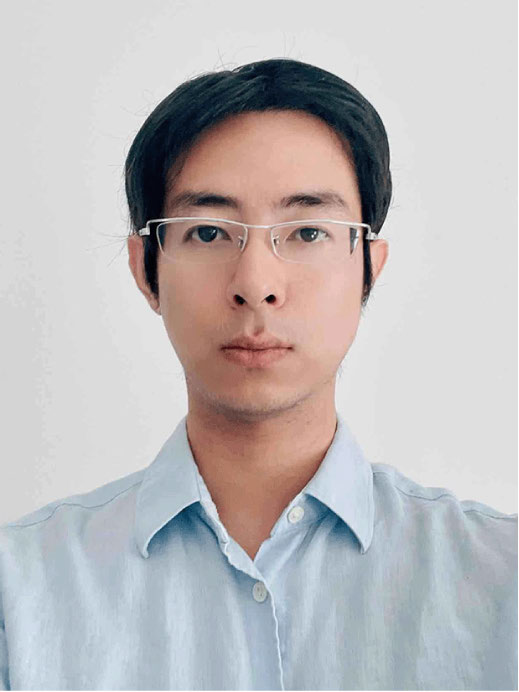}}]{Yuxuan Xia}
(Member, IEEE) received his M.Sc. in communication engineering and Ph.D. in signal and systems from Chalmers University of Technology, Gothenburg, Sweden, in 2017 and 2022, respectively. After obtaining his Ph.D., he first stayed at the Signal Processing group, Chalmers University of Technology as a postdoctoral researcher for a year, and then he was with Zenseact AB and the Division of Automatic Control, Linköping University as an Industrial Postdoctoral researcher for a year. He is currently a researcher at the School of Automation and Intelligent Sensing, Shanghai Jiaotong University. His main research interests include sensor fusion, multi-object tracking and SLAM, especially for automotive applications. He has organized tutorials on multi-object tracking at the 2020-2025 FUSION conferences and the 2024 MFI conference. He has received paper awards at 2021 FUSION and 2024 MFI.
\end{IEEEbiography}

\begin{IEEEbiography}[{\includegraphics[width=1in,height=1.25in,clip,keepaspectratio]{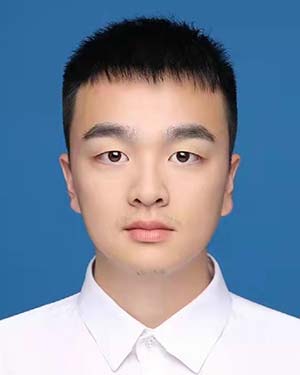}}]{Runwei Guan}
(Member, IEEE) is currently a research fellow affiliated at Thrust of AI, Hong Kong University of Science and Technology (Guangzhou). He received his PhD degree from University of Liverpool in 2024 and M.S. degree in Data Science from University of Southampton in 2021. He was also a researcher of the Alan Turing Institute and King's College London. His research interests include radar perception, multi-sensor fusion, vision-language learning, lightweight neural network, multi-task learning and statistical machine learning. He has published more than 20 papers in journals and conferences such as TIV, TITS, TMC, ESWA, RAS, Neural Networks, ICRA, IROS, ICASSP, ICME, etc. He serves as the peer reviewer of TITS, TNNLS, TIV, TCSVT, ITSC, ICRA, RAS, EAAI, MM, etc.
\end{IEEEbiography}

\begin{IEEEbiography}[{\includegraphics[width=1in,height=1.25in,clip,keepaspectratio]{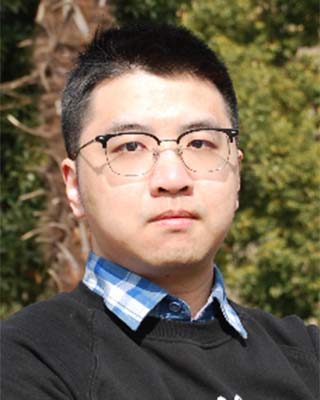}}]{Qinchen Wu}
(Student Member, IEEE) received the B.S. and M.Sc. degrees from Beihang University (BUAA), Beijing, China, in 2019 and 2021, respectively. He is currently working toward the Ph.D. degree in signal and information processing with the School of Electronic Information Engineering, Beihang University, Beijing, China. His research interests include random finite set, group target tracking, and multi-sensor data fusion.
\end{IEEEbiography}

\vfill

\begin{IEEEbiography}[{\includegraphics[width=1in,height=1.25in,clip,keepaspectratio]{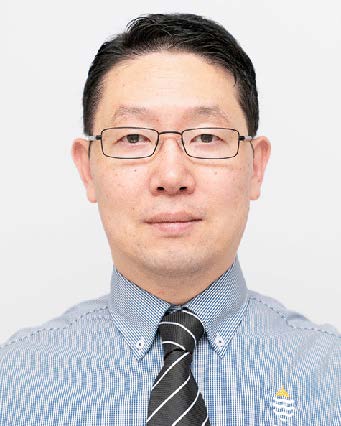}}]{Tao Huang}
(Senior Member, IEEE) received his Ph.D. in Electrical Engineering from The University of New South Wales, Australia, in 2016. Dr. Huang is a Senior Lecturer at James Cook University (JCU), Cairns, Australia. He is the Head of the International Partnerships in the College of Science and Engineering and the Intelligent Computing and Communications Lab Director. He was an Endeavour Australia Cheung Kong Research Fellow, a visiting scholar at The Chinese University of Hong Kong, a research associate at the University of New South Wales, and a postdoctoral research fellow at James Cook University. Before academia, Dr. Huang was a senior engineer, senior data scientist, project team lead, and technical lead. Dr. Huang has received the Best Paper Award from the IEEE WCNC, the IEEE Access Outstanding Associate Editor of 2023 and 2024, and the IEEE Outstanding Leadership Award. He is an Associate Editor of Scientific Reports (Nature Portfolio), IEEE Open Journal of Communications Society, IEEE Access, and IET Communications. His research interests include deep learning, intelligent sensing, computer vision, pattern recognition, wireless communications, system optimization, electronics systems, and IoT security.
\end{IEEEbiography}

\begin{IEEEbiography}[{\includegraphics[width=1in,height=1.25in,clip,keepaspectratio]{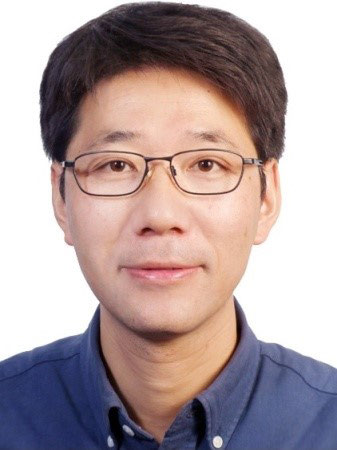}}]{Weiping Ding}
(M’16-SM’19) received the Ph.D. degree in Computer Science, Nanjing University of Aeronautics and Astronautics, Nanjing, China, in 2013. From 2014 to 2015, he was a Postdoctoral Researcher at the Brain Research Center, National Chiao Tung University, Hsinchu, Taiwan, China. In 2016, he was a Visiting Scholar at National University of Singapore, Singapore. From 2017 to 2018, he was a Visiting Professor at University of Technology Sydney, Australia. His main research directions involve deep neural networks, granular data mining, and multimodal machine learning. He ranked within the top 2\% Ranking of Scientists in the World by Stanford University (2020-2024). He has published over 360 articles, including over 170 IEEE Transactions papers. His twenty authored/co-authored papers have been selected as ESI Highly Cited Papers. He has co-authored five books. He has holds more than 50 approved invention patents, including three U.S. patents and one Australian patent. He serves as an Associate Editor/Area Editor/Editorial Board member of more than 10 international prestigious journals, such as IEEE TNNLS, IEEE TFS, IEEE/CAA Journal of Automatica Sinica, IEEE TETCI, IEEE TITS, IEEE TAI, Information Fusion, Information Sciences, Neurocomputing, Applied Soft Computing, Engineering Applications of Artificial Intelligence, Swarm and Evolutionary Computation, et al. He was/is the Leading Guest Editor of Special Issues in several prestigious journals, including IEEE TEC, IEEE TFS, Information Fusion, Information Sciences, et al. Now he is the Co-Editor-in-Chief of Journal of Artificial Intelligence and Systems, Journal of Artificial Intelligence Advances, and Sustainable Machine Intelligence Journal.
\end{IEEEbiography}
\vfill\break

\begin{IEEEbiography}[{\includegraphics[width=1in,height=1.25in,clip,keepaspectratio]{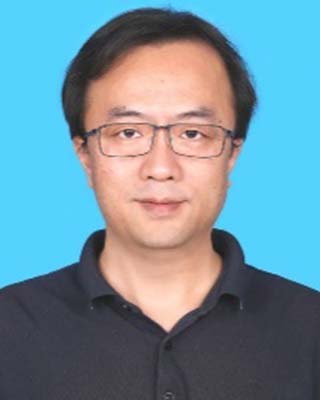}}]{Jinping Sun}
(Member, IEEE) received the M.Sc. and Ph.D. degrees from Beihang University (BUAA), Beijing, China, in 1998 and 2001, respectively. Currently, he is a Professor with the School of Electronic Information Engineering, BUAA. His research interests include statistical signal processing, high-resolution radar signal processing, target tracking, image understanding, and robust beamforming.
\end{IEEEbiography}

\begin{IEEEbiography}[{\includegraphics[width=1in,height=1.25in,clip,keepaspectratio]{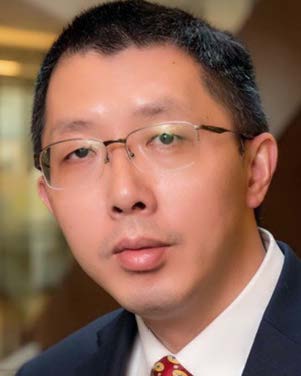}}]{Guoqiang Mao}
(Fellow, IEEE) is a Chair Professor and the Director of the Center for Smart Driving and Intelligent Transportation Systems, Southeast University. From 2014 to 2019, he was a Leading Professor, the Founding Director of the Research Institute of Smart Transportation, and the Vice-Director of the ISN State Key Laboratory, Xidian University. Before that, he was with the University of Technology Sydney and The University of Sydney. He has published 300 papers in international conferences and journals that have been cited more than 15,000 times. His H-index is 57 and was in the list of Top 2\% most-cited scientists worldwide by Stanford University in 2022, 2023, and 2024 both by Single Year and by Career Impact. His research interests include intelligent transport systems, the Internet of Things, wireless localization techniques, mobile communication systems, and applied graph theory and its applications in telecommunications. He is a fellow of AAIA and IET. He received the ``Top Editor" Award for outstanding contributions to IEEE Transactions on Vehicular Technology in 2011, 2014, and 2015. He has served as the chair, the co-chair, and a TPC member in several international conferences. He has been serving as the Vice-Director of Smart Transportation Information Engineering Society and Chinese Institute of Electronics since 2022. He was the Co-Chair of the IEEE ITS Technical Committee on Communication Networks from 2014 to 2017. He is an Editor of IEEE Transactions on Intelligent Transportation Systems (since 2018), IEEE Transactions on Wireless Communications (2014–2019), and IEEE Transactions on Vehicular Technology (2010–2020).
\end{IEEEbiography}
\vfill

\end{document}